\documentclass[journal]{IEEEtran}
\usepackage[utf8]{inputenc} 
\usepackage[T1]{fontenc}    
\usepackage{hyperref}       
\usepackage{url}            
\usepackage{booktabs}       
\usepackage{amsfonts}       
\usepackage{nicefrac}       
\usepackage{microtype}      
\usepackage{amsmath}
\usepackage{amssymb}
\usepackage{tabularx}
\usepackage{color}
\usepackage{xcolor}
\usepackage{graphicx}
\usepackage{subcaption}
\usepackage{algorithm}
\usepackage{algpseudocode}
\usepackage{siunitx}
\usepackage{multirow}
\usepackage{siunitx}
\usepackage{url}
\usepackage{bbding}

\usepackage{prettyref}
\newrefformat{fig}{Fig.~\ref{#1}}
\newrefformat{sec}{Sec.~\ref{#1}}
\newrefformat{tab}{Tab.~\ref{#1}}
\newrefformat{algo}{Algo.~\ref{#1}}
\newrefformat{eqn}{Eqn.~\ref{#1}}

\newcommand{\thickhline}{%
    \noalign {\ifnum 0=`}\fi \hrule height 1pt
    \futurelet \reserved@a \@xhline
}

\DeclareRobustCommand\onedot{\futurelet\@let@token\@onedot}
\def\onedot{\ifx\@let@token.\else.\null\fi\xspace}

\def\ie{\emph{i.e.}}

\IEEEoverridecommandlockouts                              

\title{
An Intelligent Self-driving Truck System For Highway Transportation
}

\author{Dawei Wang$^{1}$, Lingping Gao$^{2}$, Ziquan Lan$^{2}$, Wei Li$^{2\dagger}$, Jiaping Ren$^{2}$, Jiahui Zhang$^{2}$, Peng Zhang$^{2}$,\\ Pei Zhou$^{2}$, Shengao Wang$^{2}$, Jia Pan$^{1\dagger}$, Dinesh Manocha$^{3}$ and Ruigang Yang$^{2}$

\thanks{$\dagger$ denotes the corresponding author. {\tt\footnotesize jpan@cs.hku.hk, wei.li@inceptio.ai}}
\thanks{$^{1}$ Dawei Wang and Jia Pan are with Department of Computer Science, The University of Hong Kong. 
        }%
\thanks{$^{2}$ Lingping Gao, Ziquan Lan, Wei Li, Jiaping Ren, Jiahui Zhang, Peng Zhang, Pei Zhou, Shengao Wang and Ruigang Yang are with Inceptio Technology.}%
\thanks{$^{3}$ Dinesh Manocha is with Department of Computer Science, University of Maryland, College Park, USA}

}
\usepackage{cuted}
\usepackage{capt-of}
\begin{document}


\maketitle

\begin{abstract}


Recently, there have been many advances in autonomous driving society, attracting a lot of attention from academia and industry. However, existing works mainly focus on cars, extra development is still required for self-driving truck algorithms and models. In this paper, we introduce an intelligent self-driving truck system. Our presented system consists of three main components, 1) a realistic traffic simulation module for generating realistic traffic flow in testing scenarios, 2) a high-fidelity truck model which is designed and evaluated for mimicking real truck response in real world deployment, 3) an intelligent planning module with learning-based decision making algorithm and multi-mode trajectory planner, taking into account the truck's constraints, road slope changes, and the surrounding traffic flow. We provide quantitative evaluations for each component individually to demonstrate the fidelity and performance of each part. We also deploy our proposed system on a real truck and conduct real world experiments which shows our system's capacity of mitigating sim-to-real gap. Our code is available at https://github.com/InceptioResearch/IITS

\end{abstract}
\begin{IEEEkeywords}
Self-Driving, Heavy-Duty Truck
\end{IEEEkeywords}

\IEEEpeerreviewmaketitle
\section{Introduction}

Autonomous driving technology has become a billion-dollar market worldwide~\cite{viscelli2018driverless, fortune2020market}. Recently, the logistics truck with SAE level four (L4) autonomy gains more spotlights in venture capital and academia as it is believed to achieve massive production much earlier than the self-driving car. With more focused yet simpler scenarios defined in operational design domains (ODD), such as highway transportation, the self-driving truck has lower requirements for perception and prediction than the self-driving car. However, there are still several challenges that need to be resolved in the self-driving truck system including precise control with complex truck system dynamics and corresponding truck aware decision making and motion planning algorithms in highway traffic flow. In this paper, we tackle the problem of developing a practical self-driving system for a highway truck, especially focus on the planning and control (PnC) modules, which largely differ from a passenger car. 

It is well known that deep learning technique forms the cornerstone of modern autonomous driving system, which benefits the whole system from perception to localization, decision making, motion planning, and control. With regard to the PnC modules, even several learning-based algorithms, especially reinforcement learning methods~\cite{ulbrich2013probabilistic, wang2019lane, codevilla2018end, sallab2017deep} are proposed in the most recent years, they cannot seamlessly deploy to the self-driving truck system. Firstly, the truck's system dynamics are much more complex compared to a passenger car due to its low power/mass ratio, the time delay of internal engine control/air brake, prominent disturbance during gear shifting, wind, and slight road grade, etc~\cite{lu2004heavy}. The truck's dynamic model aware PnC algorithms should be developed to meet the unique requirement of trucks. Secondly, reinforcement learning-related algorithms highly depend on simulation environments, which injects a wide gap between the simulation and real world operation. A straightforward way to mitigate sim-to-real gap is system identification, which identifies the exact physical/dynamical parameters of environment relevant to the task and model it in simulation precisely. In recent years, several simulators have been developed for autonomous driving techniques, e.g., Carla~\cite{Dosovitskiy17},  AirSim~\cite{shah2018airsim}. However, these sophisticated simulators are mainly developed for cars and perception algorithms, extra development is still required for heavy-duty truck simulation.






In this paper, we introduce an intelligent self-driving autonomous truck system combining realistic traffic simulation and high-fidelity truck simulations for mitigating the sim-to-real gap. We build the system based on the service oriented middle-ware ROS2, which makes all modules decoupled and independent with each other. Then, we developed the simulation modules including a realistic truck model based on the data collected from a real truck, and construct a simulated traffic environment based on real highway roads and realistic traffic flow. On the basis of those simulation modules, we develop an intelligent planning module for trucks, including reinforcement learning-based decision maker and a multi-mode trajectory planner.

The presented system has been validated by both numerical and real world experiments. First, we conduct several experiments to validate the fidelity of our simulation modules and the results show that our simulation modules are highly close to the real world. Then, we conduct a test in the simulation environment for comparison between our proposed intelligent decision maker and rule-based decision. Finally, we deploy our system along with the pre-trained model to the real truck and demonstrate that the proposed system significantly mitigates the reality gap. In summary, the contributions of this paper are:
\begin{itemize}
    \item A complete self-driving truck system for real world logistics operation. The performance of each part, as well as the whole system, is examined with various numerical and real-world experiments. 
    \item A intelligent planning framework for self-driving truck system, covering a learning-based decision maker, multi-mode trajectory planner, increasing the system ability for interaction with complex traffic scenarios.  
    \item To tackle the challenge of sim-to-real gap and real-world deployment, we adopt system identification method to develop a realistic traffic simulation and high-fidelity truck simulation platform, which have been demonstrated their fidelity by real world experiments.
    \item During our investigation, we realize that there is no truck simulation platform that is easy to access for academia. To promote the autonomous driving truck society, our system including a high-fidelity truck model and traffic simulator is released to the public.
\end{itemize}

\section{Related Work}

\subsection{Simulation Techniques in Autonomous Driving}
Many simulators have recently been developed for autonomous driving with different focuses, such as perception realism, traffic flow and vehicle model.

\paragraph{Integrated Simulation Platform} 
Many simulators adopt computer graphics techniques to construct and render realistic environments, simulating one or more perception data channels, such as RGB-D images, LiDAR, object segmentation, etc. Popular simulators include Intel's Carla~\cite{Dosovitskiy17}, Microsoft's AirSim~\cite{shah2018airsim}, NVIDIA's Drive Constellation~\cite{nvidia-drive-constellation}, and Google/Waymo's CarCraft~\cite{waymo-carcraft}. More sophisticated simulator employs data-driven approach to render photo-realistic environments, such as Baidu's AADS~\cite{li2019aads}. However, these works mainly focus on generating realistic perception data, which are more suitable for computer vision tasks.

\paragraph{Traffic Flow Simulation}
There are simulators focusing on traffic flows, such as SUMO~\cite{behrisch2011sumo}, Vissim~\cite{fellendorf2010microscopic} and HighwayEnv~\cite{highway-env}. In particular, SUMO provides editable traffic scenarios with heterogeneous traffic agents, including road vehicles, public transport and pedestrians. Integrating SUMO and Carla, SUMMIT~\cite{cai2020summit} focuses on simulating urban driving in massive mixed traffic. HighwayEnv offers a simulator for behavioural planning in autonomous driving, which is widely used as an environment to train deep learning algorithms for high-level decision making. Although there are various works developed with HighwayEnv and SUMO, none of them has ever migrate their learned model to real world vehicle because of the huge gap between simulation and real vehicles. 

\paragraph{Truck Simulation Platforms}
Most aforementioned simulators assume the vehicle model of a passenger car. We are particularly interested in truck simulation which differs significantly from car simulation in terms of kinematics and dynamics models. Well-known truck simulators include TruckSim~\cite{trucksim} and EuroTruck~\cite{euro-truck}. Among them, EuroTruck is essentially a game with a python wrapper~\cite{europilot}, without access to the detail of its underlying truck model. On the other hand, TruckSim is a commercial software which cannot be easily accessed by the public.

In this paper, we describe the intelligent autonomous truck system, which also contains a high-fidelity autonomous simulation platform for truck, integrating simulators with various strengths, such as Carla, SUMO and TruckSim, with the aim to facilitate the development and evaluation of autonomous truck algorithms. 


\subsection{Decision Making and Planning in Autonomous Driving}
Schwarting et al.~\cite{schwarting2018planning} provide a detailed review about the schema of the decision making and planning components in autonomous driving, which divide them into three categories: sequential planning, behavior-aware planning and end-to-end planning. In our proposed system, we adopt sequential planning framework which contains a decision maker and a planner sequentially. Hence, we briefly review the previous works of these two aspects. Many existing approaches are proposed to the decision-making problems for autonomous cars.
Ulbrich et al.~\cite{ulbrich2013probabilistic} apply an online Partially Observable Markov Decision Process (POMDP) to accommodate inevitable sensor noise and make decisions in urban traffic scenarios. Wang et al.~\cite{wang2018reinforcement} present a reinforcement learning approach for lane-change maneuver. They deploy a DQN network to make decisions for lane change and with a safety guarantee. With a high-level decision, low-level planners are then used to generate feasible driving trajectories. Low-level trajectory planners for
autonomous driving trucks include polynomial curves~\cite{piazzi2002quintic}, state lattice~\cite{ferguson2008motion}, the A* family~\cite{urmson2008autonomous}, etc. However, compared to cars, trucks have more complex kinematics and challenging dynamics, which makes the existing decision making and planning system hard to directly migrate to heavy duty truck.

\section{System Overview}
\begin{figure*}
    \centering 
    \includegraphics[width=\linewidth]{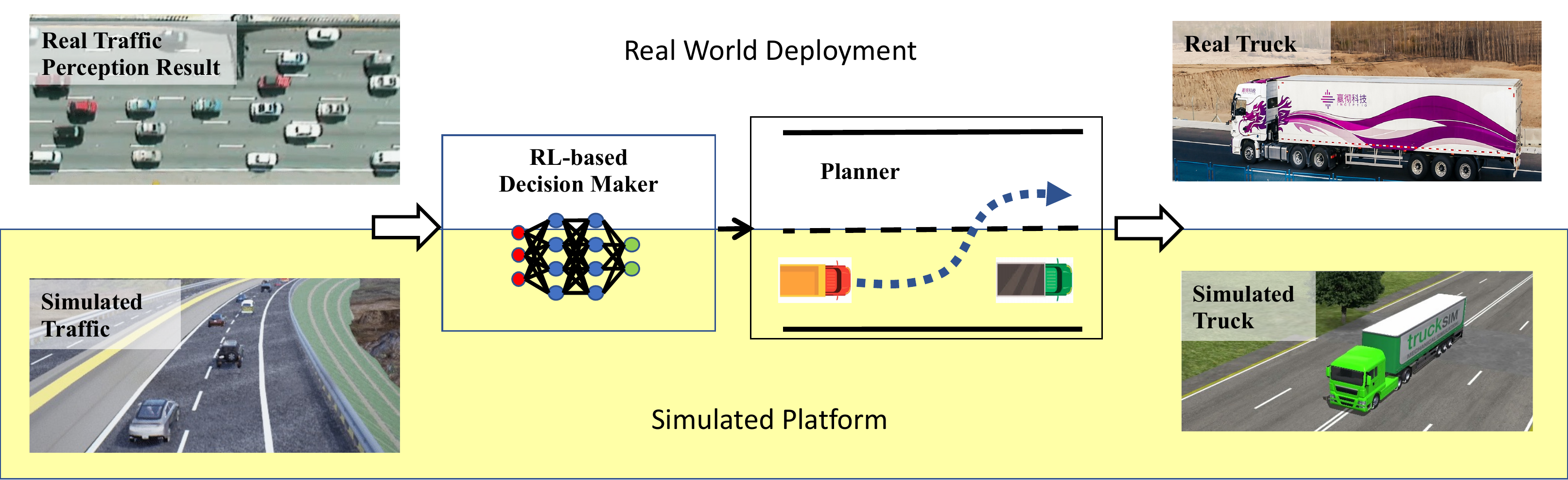}
    \caption
    {\small The overview of our proposed intelligent self-driving truck system. }
    \label{fig:deploy}%
\end{figure*}

Our proposed intelligent self-driving truck system contains three components: traffic simulation module, truck model and intelligent planning module as shown in Fig.~\ref{fig:deploy}. The traffic simulation module is designed to simulate traffic flow substituting for the perception result in real world experiment. The truck model is developed for reproducing the real truck in simulated environment precisely. The intelligent planning module consists of reinforcement learning based decision maker and a multi-mode trajectory planner, truck's constraints, road slope changes and the surrounding traffic flow.

The data flow of our proposed system can be summarized as: first, the traffic simulation module will generate realistic traffic flow interacted with ego-vehicle. Then the simulated surrounding state for ego-vehicle will be feed to a reinforcement learning based decision making module. The decision maker will output a high-level decision for ego-vehicle. After that, the planning module will conduct a feasible trajectory considering high-level decision, map information, collision avoidance and fuel efficiency. Then the control module will execute the trajectory and send the corresponding control command to the high-fidelity truck model. Finally, the truck model will output the high-fidelity response. 

The rest part of this paper is organized as follows: In Sec.~\ref{sec:traffic}, we describe the technical detail of our realistic traffic simulation first, and conduct experiment to demonstrate that our simulation can generate similar traffic which is highly close to real world data. In Sec.~\ref{sec:truck}, we describe the implement detail of our truck model and demonstrate the its fidelity. In Sec.~\ref{sec:planning}, we present the intelligent planning module including reinforcement learning based decision making, multi-mode trajectory planner and fuel efficient predictive cruise control algorithm. We conduct several numerical experiments to evaluate the decision maker and fuel-saving performance. Finally, we deploy our system to real truck and illustrate the running result in Sec.~\ref{sec:real_truck_exp}. 
\section{Realistic Traffic Simulation}
\label{sec:traffic}
\subsection{Implementation Details}
\begin{figure}[htbp] 
	\centering
    \begin{subfigure}[b]{0.30\linewidth}
        \centering
        \includegraphics[width=\linewidth]{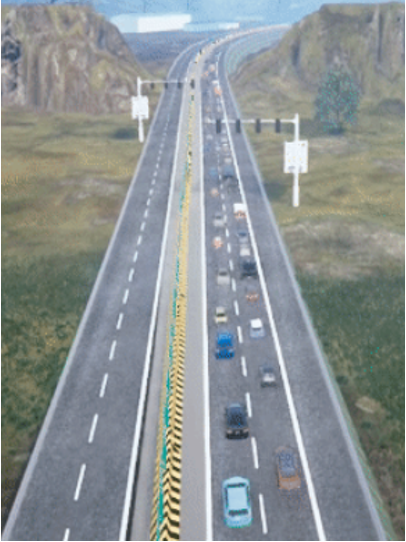}
        \caption[Network2]%
        {{\small Dense Traffic}}    
    \end{subfigure}
    \hfill
    \begin{subfigure}[b]{0.30\linewidth}  
        \centering 
        \includegraphics[width=\linewidth]{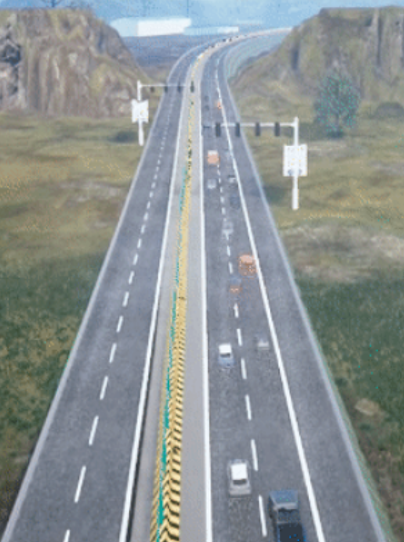}
        \caption[]%
        {{\small Medium Traffic}}    
    \end{subfigure}
    \hfill
    \begin{subfigure}[b]{0.31\linewidth}  
        \centering 
        \includegraphics[width=\linewidth]{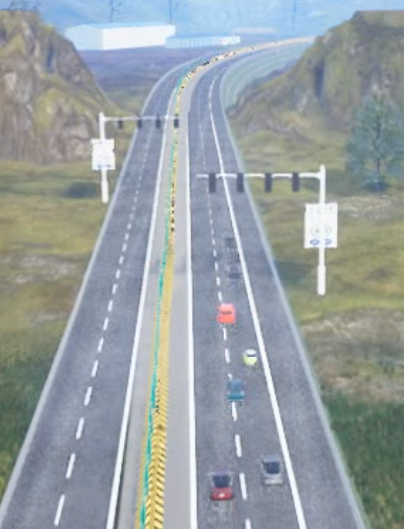}
        \caption[]%
        {{\small Sparse Traffic}}    
    \end{subfigure}
	\caption{\label{fig:trafficSimulation}%
		Results of realistic traffic simulation with different densities.
	}
\end{figure} 

We develop the traffic simulation module based on SUMO~\cite{behrisch2011sumo} to generate dynamic traffic environments for the ego-truck (as shown in Fig.\ref{fig:trafficSimulation}). The traffic simulation enhances SUMO with more a friendly Python interface for configuration and integration, and a more intelligent mode for RL training. Technically, the traffic simulation module, which is a ROS2 Python node, consists of three sub-modules: map network, traffic controller, and vehicle meta-information. The map network module describes the road connections, routes the traces, locates the vehicles, etc. The traffic controller module provides the high-level traffic control using traffic light. The vehicle meta-information module gives the attributions of vehicles, and we can adjust the vehicle behavior through this module. For different usage and  requirements, we develop four modes for realistic traffic simulation modules:

\begin{itemize}
    \item \emph{PureSim Mode.} This mode simulates the traffic flow with typical car-following model (e.g., Intelligent Driver Model (IDM)~\cite{treiber2001microsimulations}) and lane-change model (e.g., MOBIL~\cite{kesting2007general}).
    \item \emph{InterSim Mode.} This mode is built based on PureSim, and can support interactions with the ego-truck when  simulating the traffic flow. 
    \item \emph{ReSim Mode.} This mode is designed to re-simulate the traffic flow with the saved configurations. ReSim mode can guarantee the determinism of simulation and yield consistent traffic, which is needed by some learning algorithms training. 
    \item \emph{RepSim Mode.} RepSim Mode is the enhanced version of ReSim, which could not only restore the saved configurations but also the stored traffic flow,~\ie trajectories of vehicles. Powered by our unique truck trajectories and some public trajectories datasets, the simulated traffic from RepSim Mode is deterministic and with very high-fidelity.
\end{itemize}

In order to describe map information in our simulation pipeline, we adopt widely used description format, ASAM OpenDRIVE, which provides a common base for describing road networks with extensible markup language (XML) syntax, with the file extension xodr. We can support not only the maps created by hands, but also the HD map of the real road networks. With the diverse maps as the environments, we can generate various traffic flows.

\subsection{Experiment Result}
\begin{figure*}[!ht] 
	\centering
    \begin{subfigure}[b]{0.3\linewidth}
        \centering
        \includegraphics[width=\linewidth]{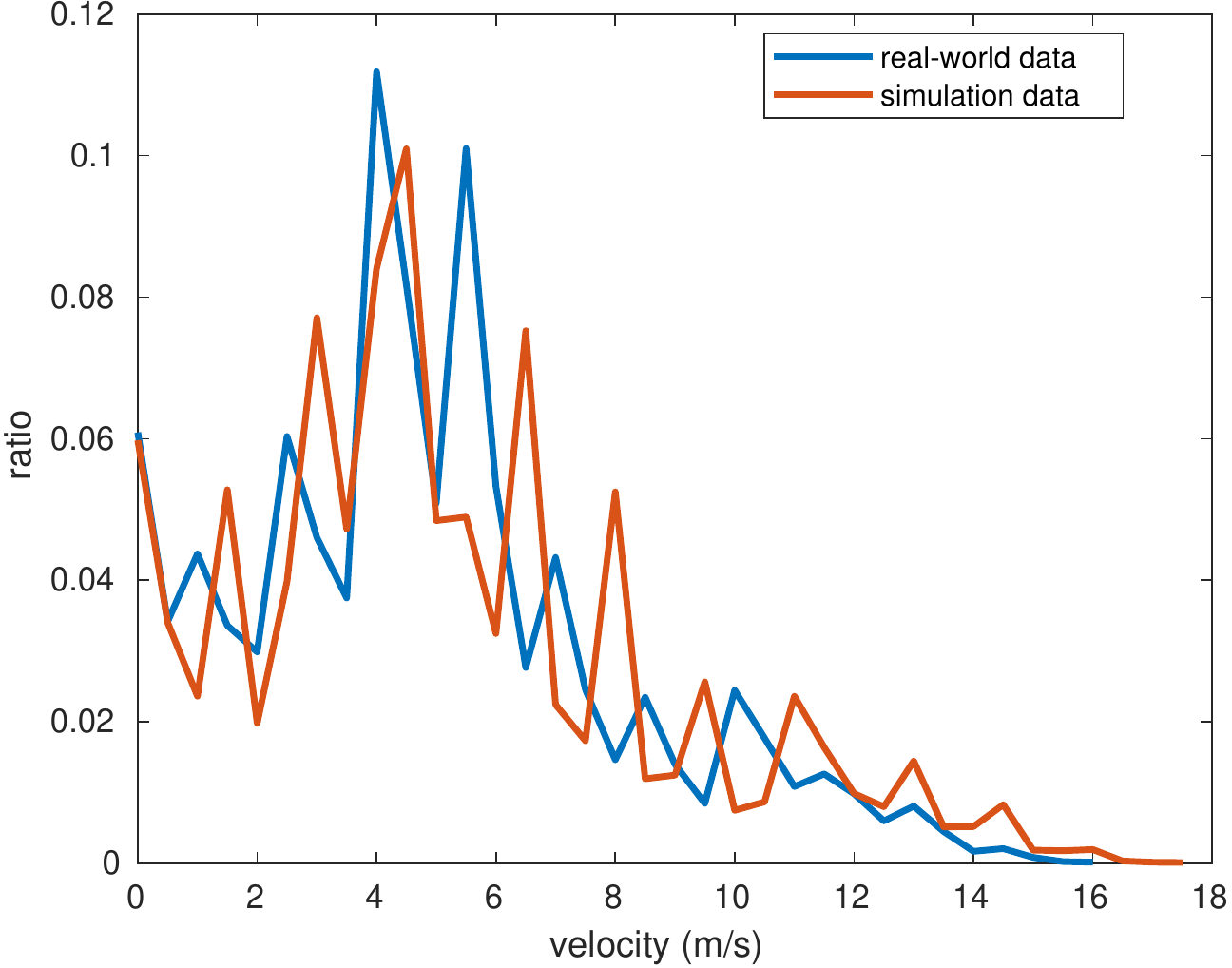}
        \caption[Network2]%
        {{\small Dense Traffic}}    
    \end{subfigure}
    \hfill
    \begin{subfigure}[b]{0.3\linewidth}  
        \centering 
        \includegraphics[width=\linewidth]{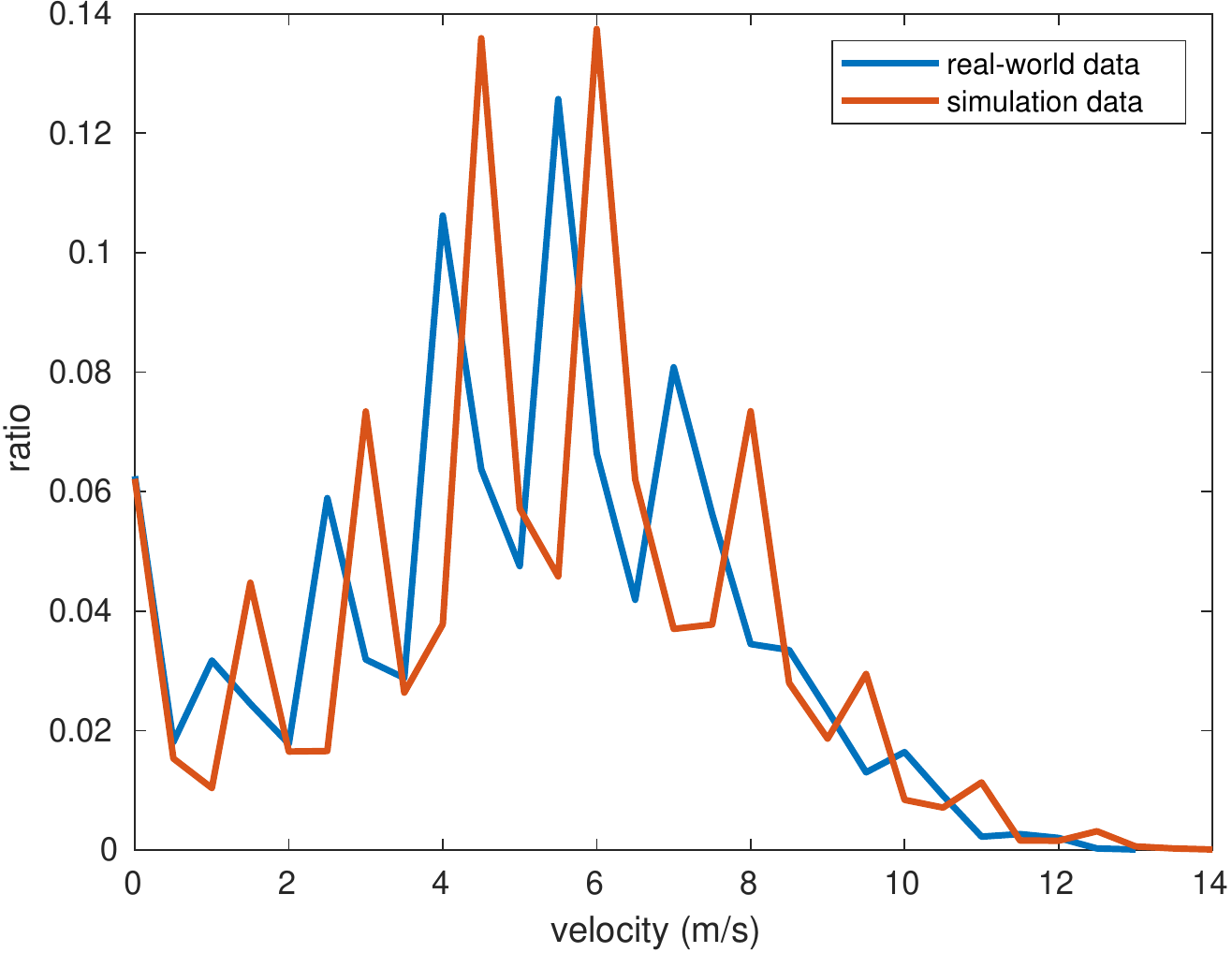}
        \caption[]%
        {{\small Medium Traffic}}    
    \end{subfigure}
    \hfill
    \begin{subfigure}[b]{0.3\linewidth}  
        \centering 
        \includegraphics[width=\linewidth]{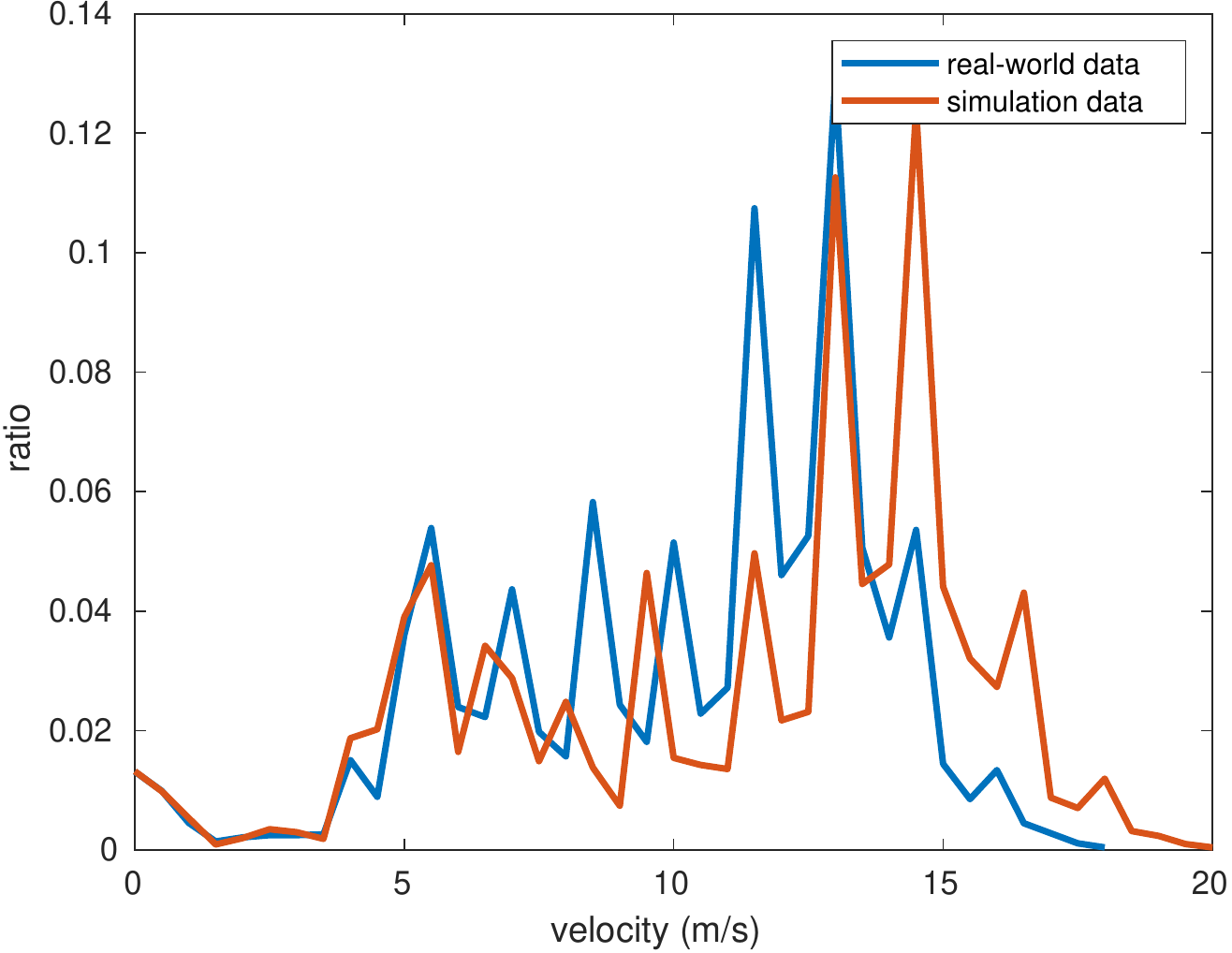}
        \caption[]%
        {{\small Sparse Traffic}}    
    \end{subfigure}
	\caption{\label{fig:vel_distri}%
		Velocity distributions of three traffic flows with different densities: (a) dense, (b) medium and (c) sparse. The blue lines indicate the distributions of real-world data. The red lines indicate the distributions of our simulation results.
	}

\end{figure*}

Currently, there are two types of evaluation for traffic simulation: user studies and statistical validations.~\cite{chao2020survey} In this paper,  we compare the velocity distributions with those of the real-world datasets, similar to~\cite{sewall2011interactive}. Specifically, We choose three datasets with different traffic flow densities (sparse, middle and dense), and the simulation result is generated under RepSim mode in system, and the real world data is collected in our test site/road in Jinan. Then, we divide our traffic flow into three levels of traffic volumes, and compare velocity distribution of real data and simulated traffic. The velocity distributions with different traffic densities are shown in Fig.~\ref{fig:vel_distri}. The result shows that our simulated traffic flow is highly close to real world data.


\section{High-Fidelity Truck Simulation Platform}
\label{sec:truck}
\subsection{Overview}
An accurate vehicle model forms the cornerstone of high-fidelity simulation. We implement our real truck model in the system to mimic the real truck experiments, because both the truck's powertrain system and kinematics are important for system deployment and also machine learning approaches training. In addition, the administrative authority has not approved self-driving truck road testing on open roads, we also have to test our methods in simulation instead of open road testing. We build our truck model based on a widely used truck simulator, Trucksim. At the same time, we re-implement the powertrain system and brake system with Simulink based on our real truck. 
\begin{figure*}[htbp] 
	\centering
	\mbox{} \hfill
	\includegraphics[width=1\linewidth]{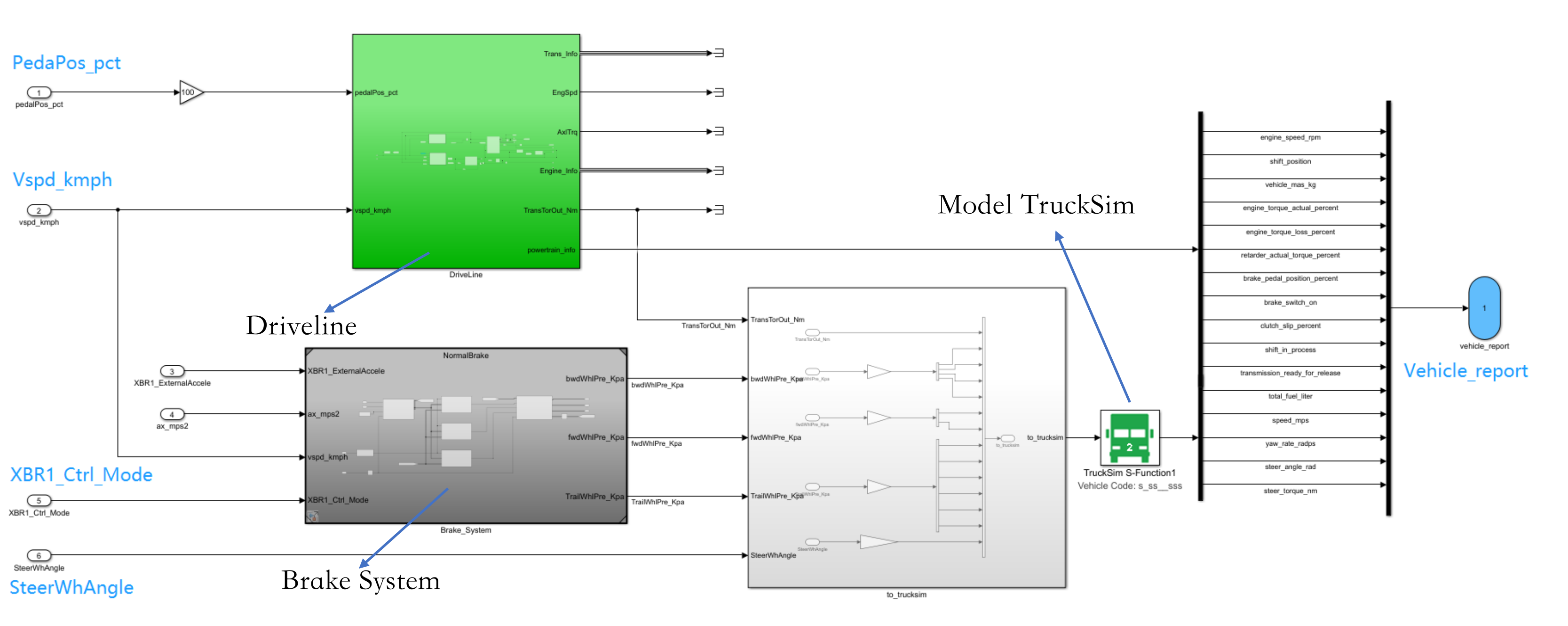}
	\hfill \mbox{}
	\vspace{-0.5cm}
	\caption{\label{fig:dynamic}%
		Overview of our truck model. The green block is the powertrain system, the dark gary block is the brake system and the while block is the interface between Trucksim and Simulink.
	}
\end{figure*} 

Details of our truck model is shown in~\prettyref{fig:dynamic}. The green block is the powertrain system, which consists engine, gearbox, clutch, engine controller and gearbox controller. The dark gray block is the brake system, consisting pneumatic brake system, brake control suspension, tire, tire brake mechanism and traction machine. The white block is the interface between Trucksim and Simulink. Trucksim is responsible for differential mechanism, vehicle body, and trailer body.

\begin{table}[]
\centering
\begin{tabular}{|c|c|c|}
\hline
Input Signal         & Metric Units & Range     \\ \hline
PedalPos             & -            & {[}0,1{]} \\ \hline
XBR1\_ExternalAccele & m/s          & -         \\ \hline
XBR1\_ctrl\_mode     & -            & 0/1       \\ \hline
SteerWhAngle         & rad          & -         \\ \hline
\end{tabular}
\caption{The input signals of our simulated truck model} 
\label{tab:input_signal}
\end{table}

The interface of our truck model complies with SAE J1939~\cite{sae1939recommended}, which is the recommended vehicle bus standard published by Society of Automotive Engineers (SAE), widely used in the heavy-truck industry. Thus, the algorithms developed on our truck model will easily meet the satisfaction of mass production on real trucks. Our truck model takes four inputs: Pedal (PedaPos\_pct), Brake Control Mode (XBR1\_ctrl\_mode), Steering Angle (SteerWhAngle) and Deceleration (XBR1\_ExternalAccele), details of each signal is shown in Tab.~\ref{tab:input_signal}. The output includes $16$ variables, including engine torque, engine speed, shift position, etc. 

\begin{figure}[htbp] 
	\centering
	\mbox{} \hfill
	\includegraphics[width=1\linewidth]{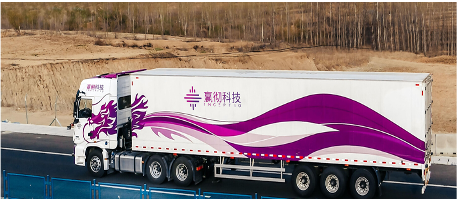}
	\hfill \mbox{}
	\caption{%
	    The prototype truck from our partner SINOTRUK, a 12 wheeler truck.
	}
	\label{fig:real_truck}
\end{figure}

The prototype of our truck model is a truck manufactured by Sinotruk Ltd, as shown in Fig.~\ref{fig:real_truck}, which is 12 wheeler heavy truck. The full load vehicle weight of our truck is 55t, and the empty load weight is 19t. The parameters and details in our model are from public data or provided by the manufacture. 

\subsection{Model Design}

\subsubsection{Kinematics}
\begin{figure}[htbp] 
	\centering
	\mbox{} \hfill
	\includegraphics[width=1\linewidth]{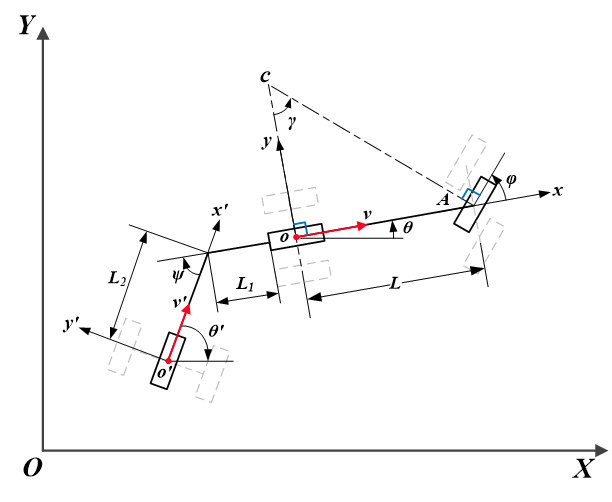}
	\hfill \mbox{}
	\caption{%
		Kinematics of vehicle-trailer system
	}
\label{fig:trailermodel}
\end{figure}

Fig.~\ref{fig:trailermodel} shows the kinematics of vehicle-trailer system, which contains two parts: a bicycle model with a unicycle model. The global reference is $\{XOY\}$, the body-fixed reference on the bicycle model is $\{xoy\}$ and the body-fixed reference on the unicycle model is $\{x'o'y'\}$. The system does not consider tire-slip angles, the vehicle velocity is $v$ and the trailer velocity is $v'$. $L_1$ refer to the tongue length and $L_2$ is the hitch length. The wheel base is $L$. $\phi, \theta, \theta'$ refer to the steering angle, vehicle heading and trailer heading. Therefore, the vehicle-trailer system model is directly given as follow:

\begin{gather}
    \Dot{x} = v*cos(\theta) \nonumber\\
    \Dot{y} = v*sin(\theta) \nonumber\\
    \Dot{\theta} = (v*tan(\phi))/L \nonumber\\
    \Dot{\phi} = \omega \nonumber\\
    \Dot{\psi} = -v*(\frac{sin(\psi)}{L_2} + \frac{L_1}{LL_2}*cos\psi tan\phi + \frac{tan\phi}{L})
\end{gather}

\subsubsection{Dynamics}

The vehicle dynamics can be first modeled as a tire-ground model via Newton's law and Lagrange's equations:
\begin{equation}  
\begin{aligned}
    F_{xf}cos\phi + F_{xr} - F_{yf}sin\phi = -F_{c}sin\beta + m\Dot{v}cos\beta \\
    F_{yf}cos\phi + F_{yf} + F_{xf}sin\phi = F_{c}cos\beta + m\Dot{v}sin\beta \\
    J\Ddot{\theta} = F_{xf}sin\phi{L_{f}} + F_{yf}cos\phi - F_{yf}L_{r}
\end{aligned}
\end{equation}

\begin{figure}[htbp] 
	\centering
	\mbox{} \hfill
	\includegraphics[width=1\linewidth]{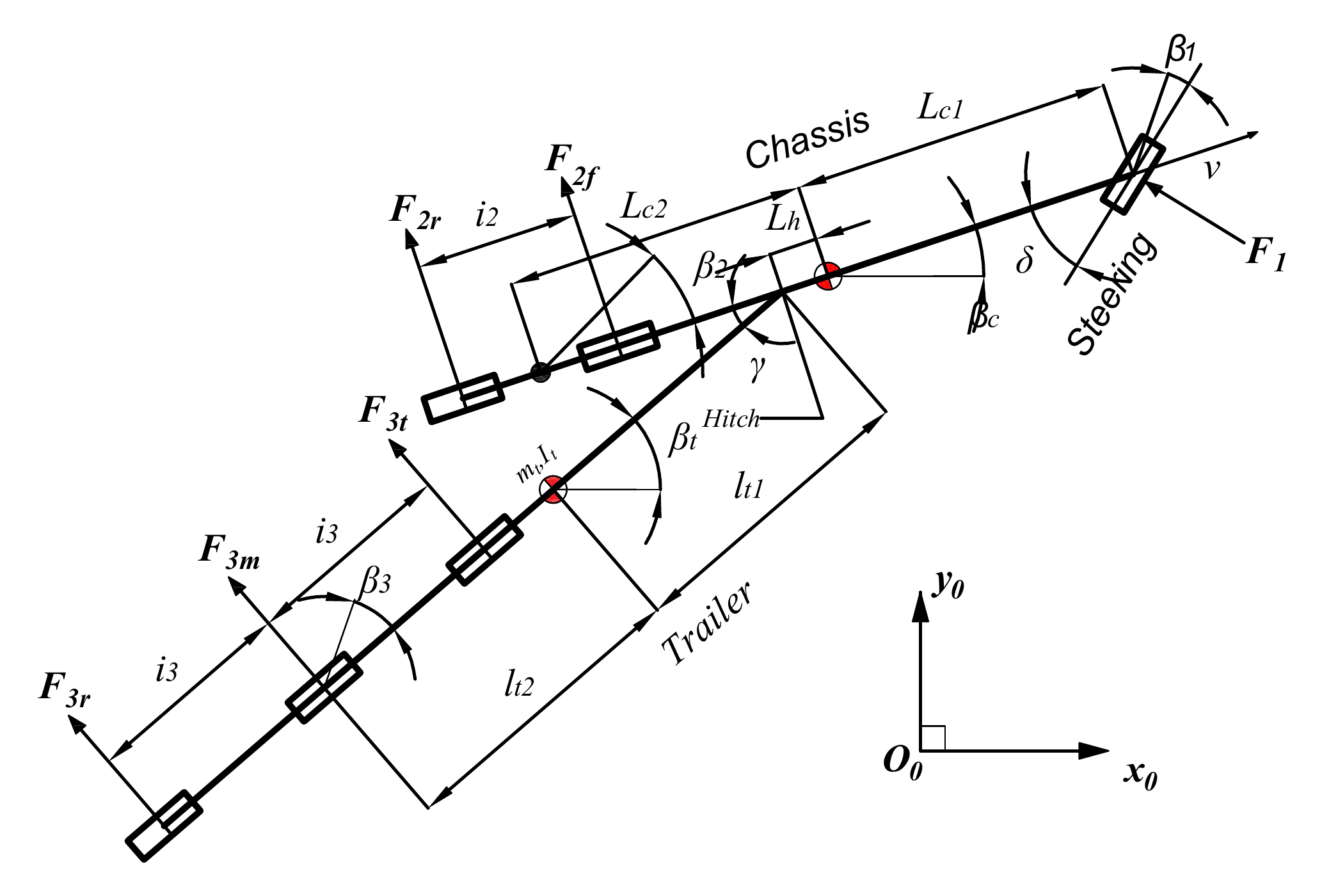}
	\hfill \mbox{}
	\caption{%
		Dynamics of truck with trailer
	}
\label{fig:trailerdynamics}
\end{figure}

Fig.~\ref{fig:trailerdynamics} illustrates the truck dynamics with trailer, which can be decomposed to two aspects: longitudinal part and lateral part.

\paragraph{Longitudinal - Drive-line \& Brake Subsystems}

Regarding the drive-line subsystem, also known as Powertrain, our intelligent truck system introduces truck engine, ECU (Engine Control Unit), AMT (Automated Manual Transmission) and TCU (Transmission Control Unit). We build those 4 modules from scratch using MATLAB Simulink using system identification method based on our truck prototype. 

\paragraph{Lateral - Dynamic Yaw-Sideslip Model}
A dynamic yaw-sideslip model shown in~\prettyref{fig:trailerdynamics} is designed to describe the truck's lateral motion in Simulink instead of the TruckSim Model. The Lagrangian mechanics is introduced to provide a governing equation for a tractor-trailer dynamic system with tandem (multiple) axles on tractor and trailer, the governing equation can be written as:
\begin{equation}
    \frac{\mathrm d}{\mathrm d t} \left( \nabla_{\dot{q}} T \right) = Q,
\label{eq:lag_eq}
\end{equation}
where $q = \left[x_c, y_c, \beta_c, \beta_t\right]^T$ is the generalized coordinates, $Q$ is the generalized force and $T$ represents the total kinetic energy.

For the energy term on the left-hand side of Eq. \ref{eq:lag_eq}, the combined vehicle's total kinetic energy can be written as:
\begin{equation}
    T = \frac{I_c \,{\omega_c }^2 }{2}+\frac{I_t \,{\omega_t }^2 }{2}+\frac{m_c \,{\left(v^2 +{v_{\textrm{yc}} }^2 \right)}}{2}+\frac{m_t \,{\left(v^2 +{v_{\textrm{yt}} }^2 \right)}}{2},
\end{equation}
where $I$ stands for rotational inertia, $\omega$ stands for yaw rate, $m$ stands for mass, $v$ represents the longitudinal speed, and the additional subscript $c$ and $t$ are used to indicate the tractor (chassis) and the trailer respectively. The kinematics of the tractor-trailer system also stipulates:
\begin{equation}
    v_{\textrm{yt}} = v_{\textrm{yc}} -L_h \,\omega_c -L_{\textrm{t1}} \,\omega_t.
\end{equation}
Then the gradient of energy can be written as:
\begin{equation}
    \begin{aligned}
        \nabla_{\dot{q}} T =\qquad \qquad \qquad \qquad \qquad\qquad&\\ 
    \left(\begin{array}{cccc}
    m_c +m_t  & 0 & 0 & 0\\
    0 & m_c +m_t  & -L_h \,m_t  & -L_{\textrm{t1}} \,m_t \\
    0 & -L_h \,m_t  & m_t \,{L_h }^2 +I_c  & L_h \,L_{\textrm{t1}} \,m_t \\
    0 & -L_{\textrm{t1}} \,m_t  & L_h \,L_{\textrm{t1}} \,m_t  & m_t \,{L_{\textrm{t1}} }^2 +I_t 
    \end{array}\right) 
    &\dot{q},
    \end{aligned}
    \label{eq: grad_energy}
\end{equation}
where $\dot{q} = \left[v, v_\textrm{yc}, \omega_c, \omega_t\right]^T$.

For the force term on the right-hand side of Eq. \ref{eq:lag_eq}, The virtual work done by DOF perturbations can be written as:
\begin{equation}
\begin{array}{ll}
&\Delta W = (-F_1\delta - (F_{3f} + F_{3m} + F_{3r})\Delta\beta_t)\Delta x_c \\
&+F_1\sqrt{1-\delta^2}(\Delta y_c + L_{c1}\Delta\beta_c) \\
& +F_{2f}(\Delta y_c - (L_{c2} - i_2)\Delta\beta_c) \\
& +F_{2r}(\Delta y_c - (L_{c2} + i_2)\Delta\beta_c) \\
& +F_{3f}(\Delta y_t - (L_{t2} - i_3)\Delta\beta_t) \\
& +F_{3r}(\Delta y_t - (L_{t2} + i_3)\Delta\beta_t) \\
& +F_{3m}(\Delta y_t - L_{t2}\Delta\beta_t)  + 
    m_c \, \omega_c \, v \, \Delta y_c + 
    m_t \, \omega_t \, v \, \Delta y_t,
\end{array}
\end{equation}
where $y_t = y_c - \beta_c \, L_h - \beta_t \, L_{t1}$ is complying with the kinematic constraints. In order to obtain a linear system, coefficient $\sqrt{1-\delta^2}$ is approximated as 1. The generalized force term $Q$ can be represented as the gradient vector of the virtual work with respect to the virtual displacements along each DOF ($\Delta q$): $Q = \nabla_{\Delta q} (\Delta W) $.

By defining the system states as $s = \left[v_\textrm{yc}, \omega_c, \omega_t, \gamma \right]^T$ and front-wheel steer angle input $\delta$, the dynamics of the states can be obtained as:
\begin{equation}
    H \dot{s} = A_h s + B_h \delta,
\end{equation}
in which $A_h$, $B_h$ are calculated from differentiating $Q$ with respect to $s$, and $H$ is derived from reducing the kinetic potential in Eq. \ref{eq: grad_energy}:
\begin{equation}
H = 
\left(\begin{array}{cccc}
m_c +m_t  & -L_h \,m_t  & -L_{\textrm{t1}} \,m_t  & 0\\
-L_h \,m_t  & m_t \,{L_h }^2 +I_c  & L_h \,L_{\textrm{t1}} \,m_t  & 0\\
-L_{\textrm{t1}} \,m_t  & L_h \,L_{\textrm{t1}} \,m_t  & m_t \,{L_{\textrm{t1}} }^2 +I_t  & 0\\
0 & 0 & 0 & 1
\end{array}\right).
\end{equation}
Finally the state space model in the formal form is:
\begin{equation}
    \dot{s} = A s + B \delta,
\end{equation}
where $A = H^{-1} A_h$ and $B = H^{-1} B_h$.

\subsection{Fidelity Evaluation}
\begin{figure*}[!ht] 
	\centering
    \begin{subfigure}[b]{0.49\linewidth}
        \centering
        \includegraphics[width=\linewidth]{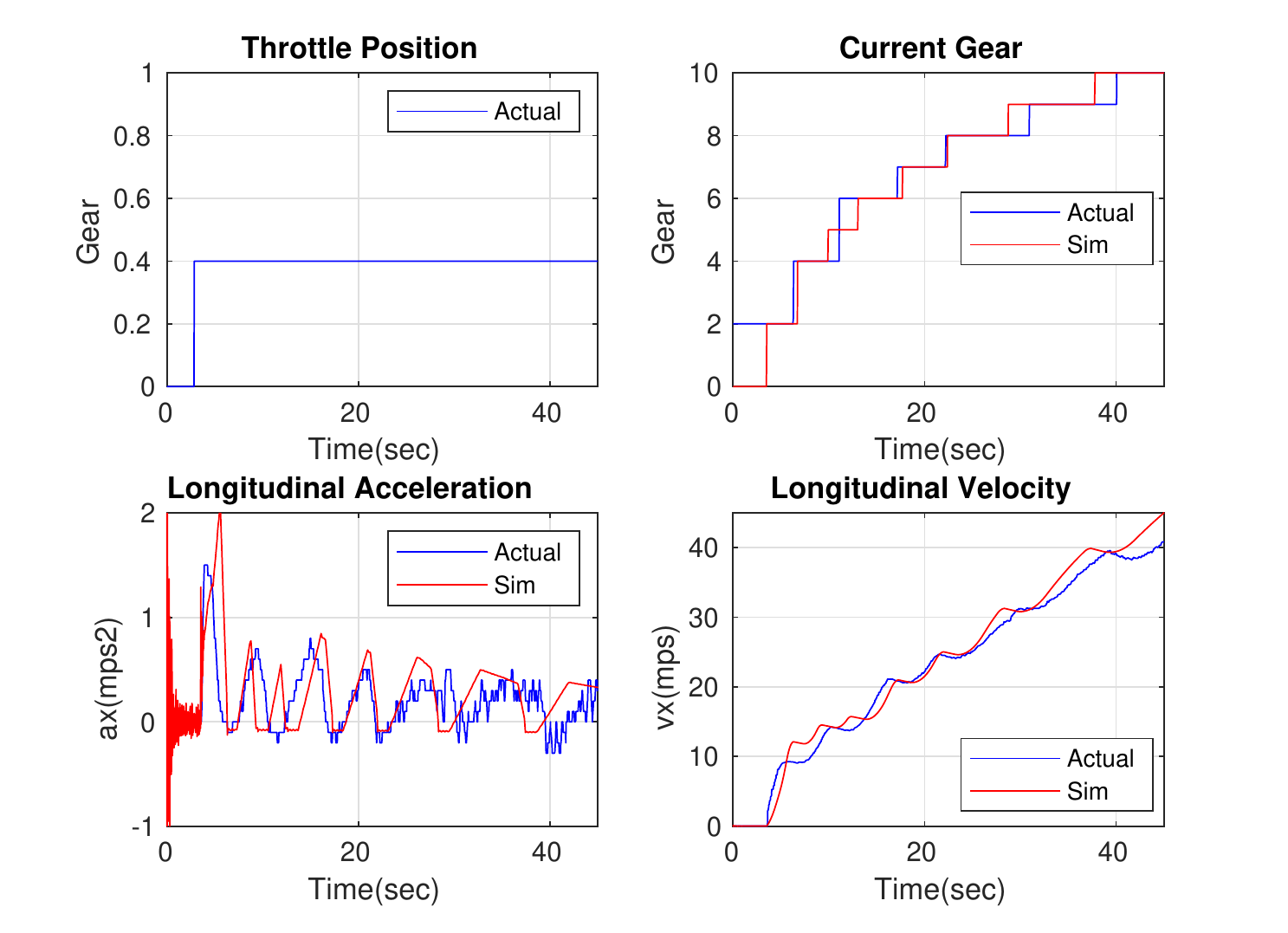}
        \caption{\label{subfig:driveline_a} Drive-line}
    \end{subfigure}
    \hfill
    \begin{subfigure}[b]{0.49\linewidth}  
        \centering 
        \includegraphics[width=\linewidth]{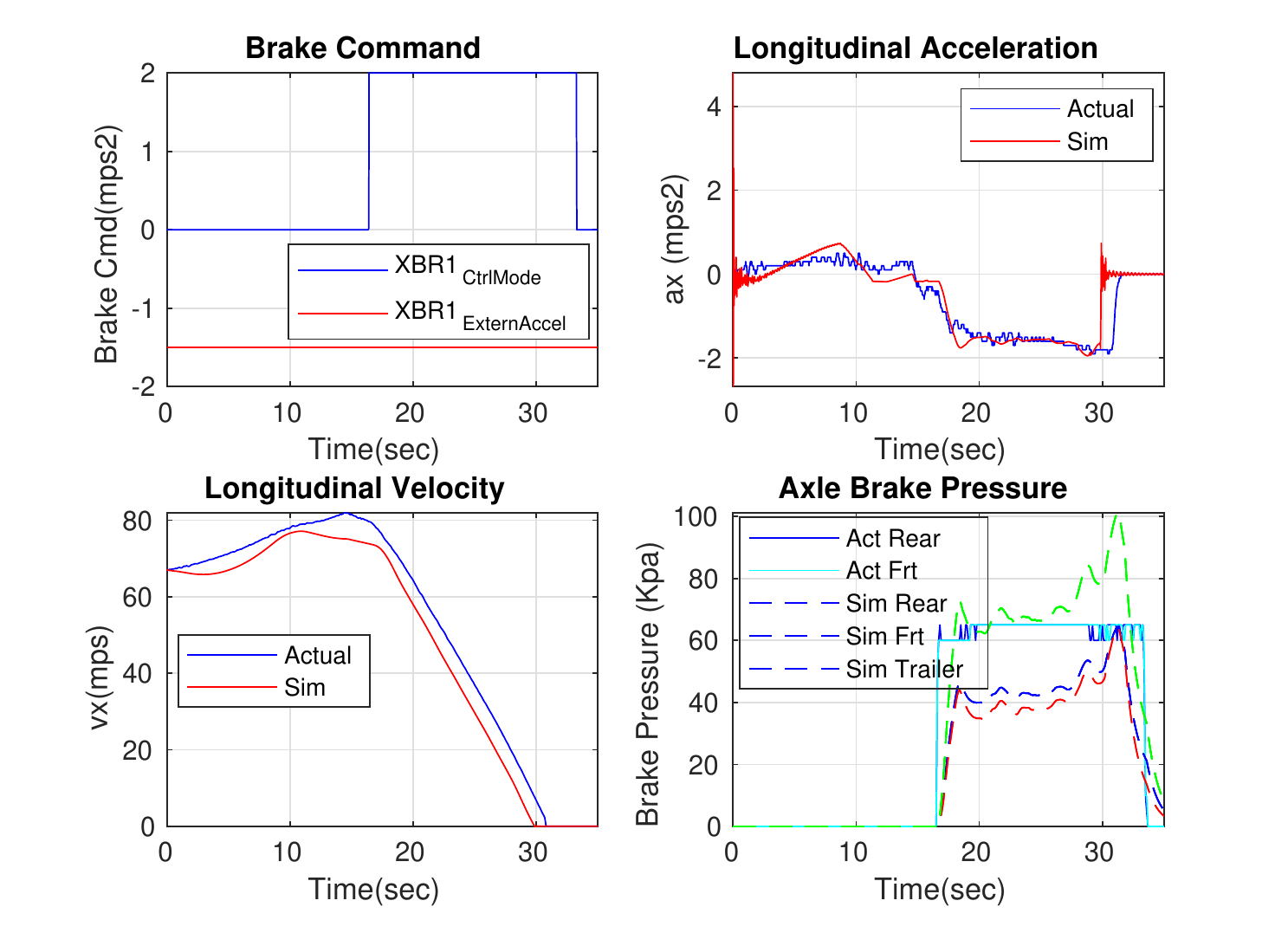}
        \caption{\label{subfig:driveline_b} Brake}
    \end{subfigure}
	\caption{\label{fig:driveline}%
		Drive-line and brake subsystems experiments. Red lines indicate simulated results, and blue lines are real data from the truck. The results demonstrate that our longitudinal model is highly close to the real truck.
	}
\end{figure*} 

\subsubsection{Fidelity of Longitudinal Model}
To verify the accuracy of the longitudinal model, we collect some data from a real truck running with a human driver for ten times, then replay the truck control commands in our simulator and compare the difference in vehicle response. ~\prettyref{fig:driveline} shows some representative results of the experiments on verifying the drive-line and brake subsystems, with Fig.~\ref{subfig:driveline_a} and Fig.~\ref{subfig:driveline_b} corresponding to the truck's acceleration and brake motions respectively, where the red lines indicate the simulated results and the blue lines are the real data from trucks. For acceleration test, we keep the throttle position at 40\% for 40 sec, and collect the vehicle status about gear (0-10), longitudinal acceleration ($m/s^2$) and velocity ($m/s$) from real vehicle and simulation module simultaneously. The result shows in Fig.~\ref{subfig:driveline_a}, the red lines are very close to the blue lines indicate our simulated model's response is highly close to the real truck. 
For brake test, we keep the throttle at 0 all the time and activate the brake at 17 sec (CtrlMode means brake activation status, 0 indicates deactivated, 2 indicates brake activated. ExternAccel means deceleration, we keep it at $-1.5m/s^2$ during our test), and collect the longitudinal velocity ($m/s$), acceleration ($m/s^2$) and brake pressure(Kpa). The result shown in Fig.~\ref{subfig:driveline_b}, the red lines are close to their corresponding blue one, demonstrate that our model's brake system is close to the real truck. Overall, the longitudinal velocity accuracy of all experiments is $89\%$.

\begin{figure}[!h] 
	\centering
        \includegraphics[width=\linewidth]{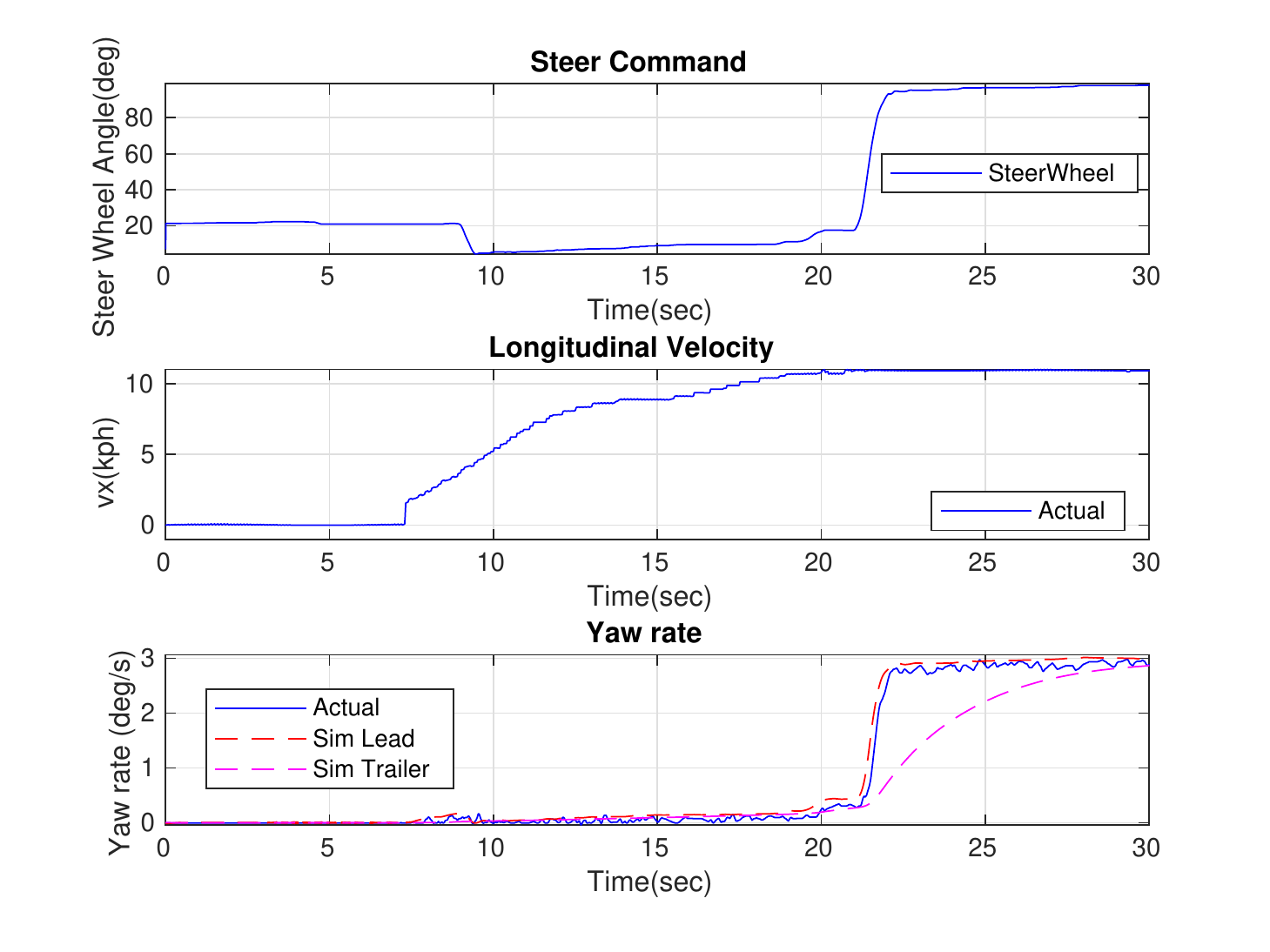}
	\caption{\label{fig:exp-steering-system}%
		Steering system experiments. Red lines indicate simulated results, and blue lines are real data from the truck. The results demonstrate that our TruckSim lateral model is highly close to the real truck.
	}
\end{figure} 
\subsubsection{Fidelity of Lateral Model}
In this paper, we adopt two lateral dynamic models, one is TruckSim model which relies on a commercial license, another one is our proposed yaw-sideslip model, which will be released in our open-source system. The Trucksim model combines steering system, solid axles and tire system together so that the physical characteristics of the real truck can be restored to the utmost extent. Same as the evaluation experiment of longitudinal model, we replay the control commands in our simulator and compare the responses. Results in Fig.~\ref{fig:exp-steering-system}, the input command of steer is shown in the top figure and the longitudinal velocity during the testing is presented in the middle. We collect the yaw rate of the tractor (degree/s), the blue line indicates the actual yaw rate of real truck, and the red line indicates the yaw rate response in our simulated model. The result shows that our truck model (TruckSim)'s lateral response is highly close to real truck experiment, and the overall lateral yaw rate steady accuracy is 92\% in our experiments. 

\begin{figure}[!h] 
	\centering
    \includegraphics[width=\linewidth]{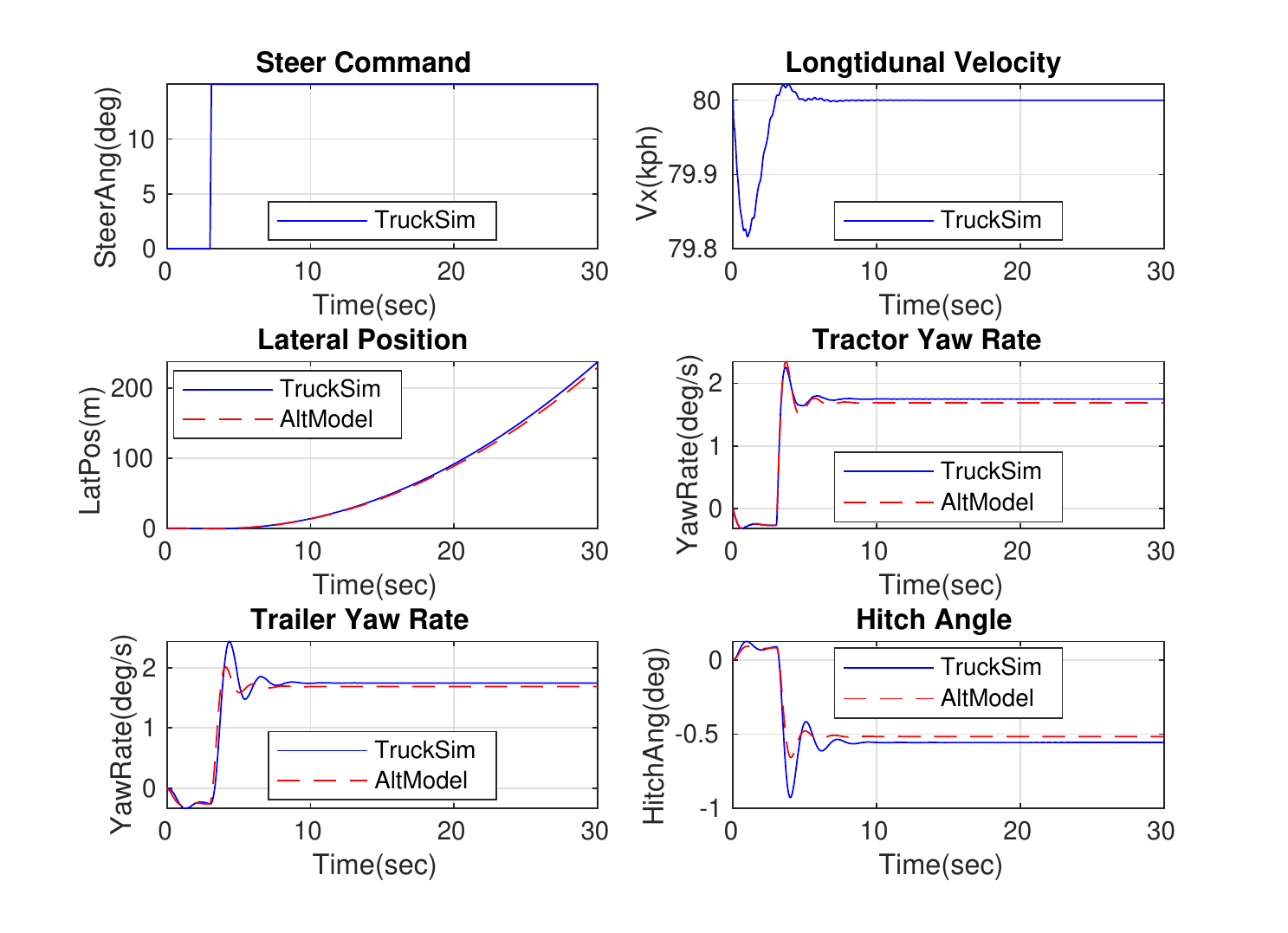}
	\caption{\label{fig:step_comb}%
		Results of the experiments on verifying the accuracy of the dynamic yaw-sideslip model. Red lines indicate results of alternative model, and blue lines are data from TruckSim model. The results demonstrate that our dynamic yaw-sideslip model is highly close to TruckSim model.
	}
\end{figure} 

\subsubsection{Fidelity of Dynamic Yaw-sideslip Model}
Since the TruckSim is a commercial, closed source software, we designed the dynamic yaw-sideslip model in Matlab Simulink as an alternative choice of TruckSim model, thus we conduct an experiment to compare the dynamic yaw-sideslip model with the TruckSim lateral model. During the comparison experiment, we set the truck speed at \SI{80}{km/h} and we fed a step steering command of 15 degrees to each model. Results are shown in  Fig.~\ref{fig:step_comb}, we collect the yaw rate response of tractor and trailer, the lateral position of the tractor and the hitch angle between the tractor and trailer, the blue lines show the result of the simulation results from TruckSim lateral model, and the red lines indicate the results from our proposed dynamic yaw-sideslip model. The result shows that our proposed yaw-sideslip lateral model is highly close to our TruckSim lateral model, which has already been demonstrated its fidelity in the previous section. In total, the overall accuracy of these experiments is 96\%.

\section{Intelligent Decision and Planning}
\label{sec:planning}
In this section, we will describe the reinforcement learning based decision making firstly, then we will present the technical detail of the planning module in our system, finally, we will conduct several experiments to evaluate our proposed intelligent decision making module. 
\begin{figure}[htbp] 
	\centering
	\mbox{} \hfill
	\includegraphics[width=1\linewidth]{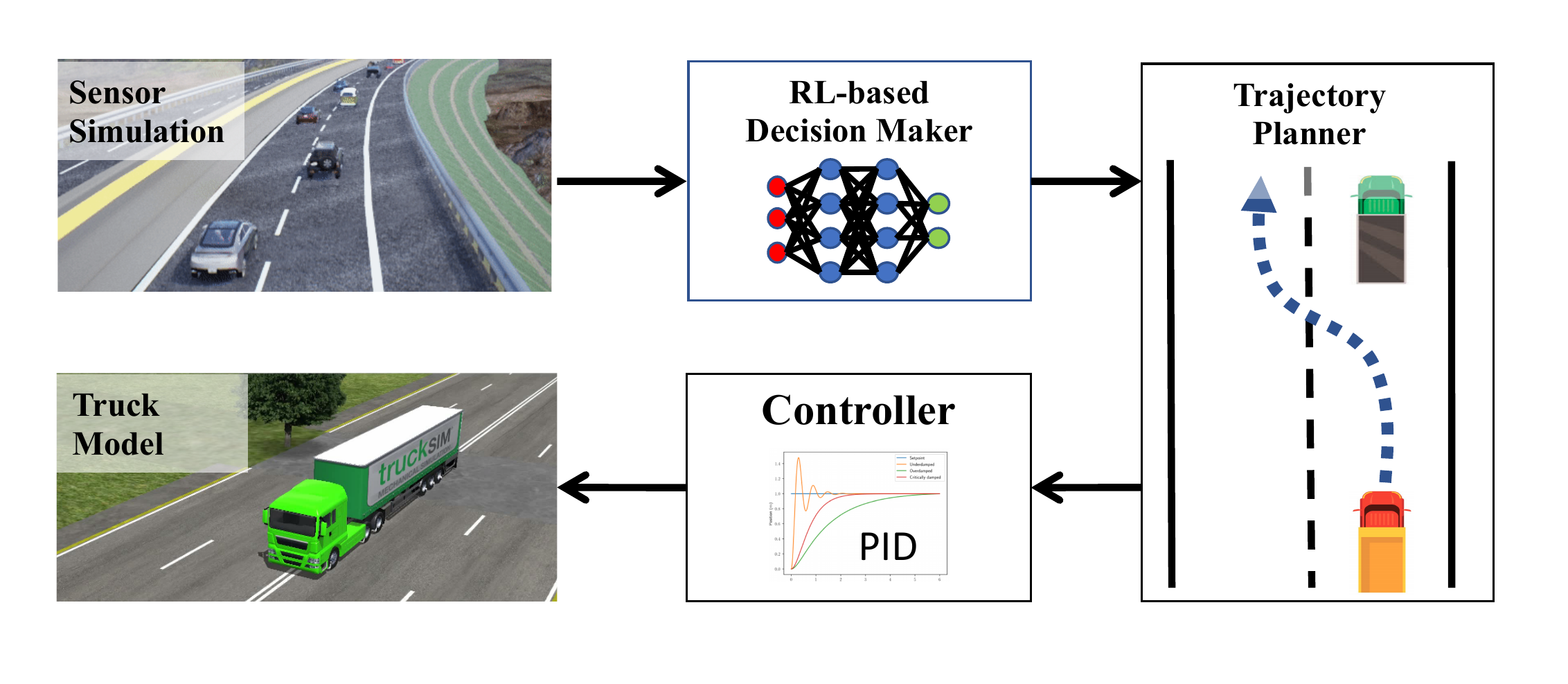}
	\hfill \mbox{}
	\caption{%
	    The pipeline of our proposed intelligent autonomous truck system. Specifically, the traffic simulation will generate realistic traffic flows interacted with ego-vehicle, then the intelligent planning module (decision, planning and control) will leverage the traffic information to generate  the control command for the ego-truck, finally the truck model will take the control command and output high-fidelity response. 
	}
\label{fig:rl-pipeline}
\end{figure}

\subsection{Reinforcement Learning Based Decision Making}
In this section, we introduce our reinforcement learning based decision maker, including problem formulation and network design. 

\subsubsection{Problem Formulation}
 The self-driving trucks evaluate and improve its decision-making policy by interacting with the environment including surrounding vehicles and lanes in a trial-and-error manner. This process can be formulated as a sequential decision-making problem, which can be solved using a reinforcement learning framework~\cite{van2015deep}. In the RL settings, the problem is formulated as a Markov Decision Process (MDP), which is composed of a five-tuple $(\mathcal{S},\mathcal{A},r({s}_t,{a}_t),\mathcal{P}({s}_{t+1}\mid {s}_t,{a}_t),\gamma)$. At time step $t$, the agent selects the action ${a}_t \in \mathcal{A}$ by following a policy $\pi$ in the current state. The agent is transferred to the next state ${s}_{t+1}$ with the probability $\mathcal{P}({s}_{t+1}\mid {s}_t,{a}_t)$ after executing ${a}_t$. Additionally, the environment returns a reward signal $r(s_t, a_t)$ to describe whether the underlying action ${a}_t$ is good for reaching the goal or not. For brevity, we rewrite it as $r_t = r({s}_t,{a}_t)$. By repeating this process, the agent interacts with the environment and obtains a trajectory $\tau = {s}_1,{a}_1,{r}_1,\cdots,{s}_T,{a}_T,{r}_T$ at the terminal time step $T$. The discount cumulative reward from time step $t$ can be formulated as ${R}_t = {\sum}^T_{k=t}{\gamma}^{k-t}r_k$, where $\gamma\in (0,1)$ is the discount rate that determines the importance of future rewards. The goal of RL is to learn an optimal policy $\pi^*$ that can maximize the expected overall discounted reward:
\begin{equation}
\begin{aligned}
&\mathcal{\pi^*} = \mathbb{\arg\max_{\pi}}E_{s,a \sim \pi,r}\left[R_1\right].
\end{aligned}
\end{equation}
Typically, two kinds of value functions are used to estimate the expected cumulative reward for a specific state:
\begin{equation}
\begin{aligned}
&\mathbf{V^\pi(s)} = \mathbb{E_{\pi}}\left[R_1 | s_1 = s\right],
\end{aligned}
\end{equation}
\begin{equation}
\begin{aligned}
\mathbf{Q^\pi(s,a)} = \mathbb{E_{\pi}}\left[R_{1}|s_1 = s,a_1 = a\right].
\end{aligned}
\end{equation}
To improve the robustness of the lane-change decision-maker based on reinforcement learning while reducing the difficulty of training, we discretize the action space of the lane-change problem, and use the double DQN algorithm~\cite{van2015deep} to solve it. In reference to both double Q-learning~\cite{hasselt2010double} and DQN~\cite{mnih2015human}, double DQN propose to evaluate the greedy policy according to the online network, but using the target network to estimate its value. Its update is the same as for DQN, but replacing the target $Y_t^{DQN}$ with
\begin{equation}
\begin{aligned}
\mathbf{Y}_t = R_{t+1} + \gamma Q\left(S_{t+1},\arg\max_aQ\left(S_{t+1},a,\theta_t\right),\theta_t^-\right).
\end{aligned}
\label{eqn:target_y_t}
\end{equation}
Double DQN replaces the weight of the second network $\theta_t^{'}$ with the weights of the target network $\theta_t^-$ for the evaluation of the greedy policy in comparison to double Q-learning. For the update method of the target network, double DQN still adopts DQN method, and remains a periodic copy of the online network.
\begin{figure}[htbp] 
	\centering
	\mbox{} \hfill
	\includegraphics[clip, trim=0 20 0 0, width=1\linewidth]{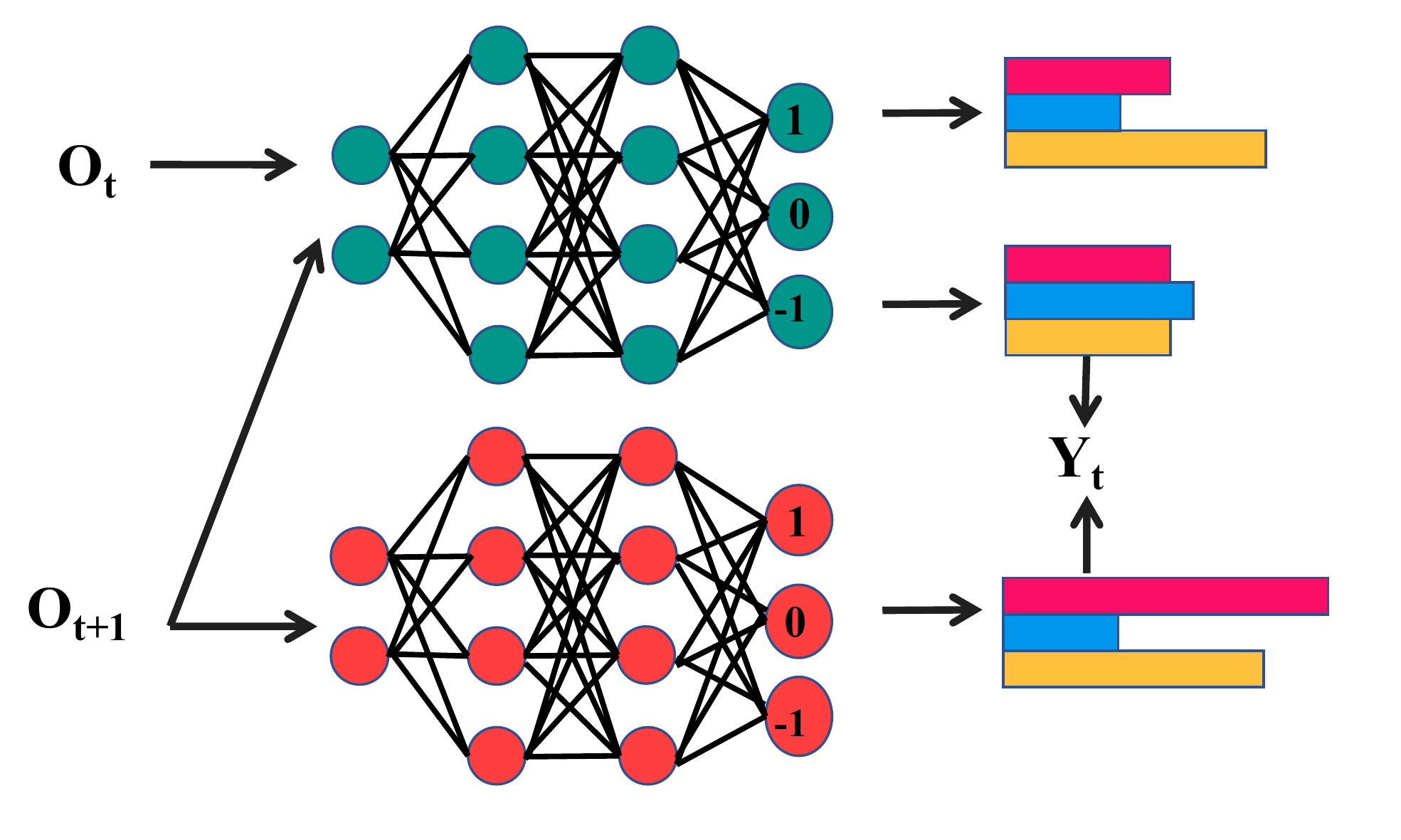}
	\hfill \mbox{}
	\caption{\label{fig:networkstructure}%
		Reinforcement learning based decision-making neural network's architecture and update method.
	}
\end{figure} 
\subsubsection{State Space}
In our formulation, the state $s$ is defined as following: 
\begin{equation}
\begin{aligned}
\mathbf{s} = \langle \mathbf{p}, \mathbf{v}, \mathbf{v}_t \rangle + \sum_{i}^N \langle ^i{o}_p^t, ^i{o}_v^t\rangle,
\end{aligned}
\label{eqn:state}
\end{equation}
where $p$ and $v$ indicate the location and velocity of ego-car, $v_t$ is the speed profile in the next three seconds, $N$ is the number of the observable neighbors, ${o}_p^t$ and ${o}_v^t$ are relative position and velocity of neighbors respectively.
\subsubsection{Action Space}
For the action $a$ of the agent, we use ${R}^3$ vectors for our action space, which includes: change left ($-1$), straight forward ($0$) and change right ($1$).
\subsubsection{Reward function}
As a key element of the RL framework, the reward drives the agent to reach the goal by rewarding good actions and penalizing poor actions. For a lane change process, safety and efficiency are the main concern. Therefore, our objective is to achieve the reference speed profile with as few lane changes as possible while ensuring safety.

The process of change lanes will not only increase the probability of danger, but also reduce the efficiency of transportation. To avoid the meaningless lane change behavior, we give a penalizing reward $r_{ch}$ when a lane change decision is made. In our simulator, we limit the self-driving trucks to only drive in two lanes of the road. That means when a decision to change lanes to the left is made while the truck is on the left lane, it stays in the current lane, but a penalizing reward $r_{ch}$ will also be given.

For efficiency, self-driving should try to meet the requirements of driving at a reference speed generated by the planner. To do so, we define the following reward according to the speed of the truck:
\begin{equation}
\begin{aligned}
\mathbf{r}_v = \lambda\left|v - v_{ref}\right|,
\end{aligned}
\label{eqn:state}
\end{equation}
where $v$ denotes the car’s current speed, and $v_{ref}$ is reference speed planned by speed planner last clock cycle while $\lambda$ is a normalizing coefficient.

For safety, we hope that the truck can change lanes while ensuring safety. Therefore, we will use the rule-based method to determine whether the lane change decision at the current moment is dangerous according to the observation. If the truck makes a lane change decision at a dangerous moment, we will give a larger penalty $r_{sa}$.

In general, our reward function goes as:
\begin{equation}
\begin{aligned}
\mathbf{r} =r_{ch} + \lambda\left|v - v_{ref}\right| + r_{sa}.
\end{aligned}
\label{eqn:state}
\end{equation}
Considering the safety and efficiency of truck transportation,we set $r_{ch} = -10$, $\lambda = 1$, $r_{sa} = -20$. 

\subsubsection{Network Design}
In the DRL network, we take four consecutive observations in the past three seconds ${S}_{t-3}$, ${S}_{t-2}$, ${S}_{t-1}$, ${S}_t$ as input. Such input can enable our agents to infer the motion of surrounding cars, and thus make more reasonable lane change decisions. Because the method of discretizing the action space reduces the difficulty of the decision-making problem, we only use a few fully connected layers to build policy network. The input and output layers have 216 and 3 neurons respectively, while the total number of neurons in the hidden is (256, 512, 256). The architecture consists of two networks with the same structure as shown in Fig.~\ref{fig:networkstructure}: the value network (green) for select action and the target network (red) for evaluating the value of the underlying state. The online network's input are $O_t = [{S}_{t-3}, {S}_{t-2}, {S}_{t-1}, {S}_t]$ and $O_{t+1} = [{S}_{t-2}, {S}_{t-1}, {S}_t, {S}_{t+1}]$ to predict value of the current state and action of the next state. The target network's input is $O_{t+1} = [{S}_{t-2}, {S}_{t-1}, {S}_t,{S}_{t+1}]$ to predict target value.

\subsection{Trajectory Planning Module}
To execute high-level decisions, we deploy a trajectory planning module, which can be divided into two modes: lane change mode and lane keeping mode. In addition, the reference speed profile for the trajectory is planned by a fuel efficient speed planner, which we adopt predictive cruise control (PCC) algorithm~\cite{lattemann2004predictive}.

\subsubsection{Fuel Efficient Speed Planner}
\label{sec:fuel_saving}
Since the fuel consumption achieves 30\% of total operation cost in logistic industry, fuel efficiency in autonomous truck system becomes more and more important currently. Predictive Cruise Control algorithm~\cite{lattemann2004predictive} is widely adopted fuel saving method for heavy duty truck, which leverage the road slope change in front of the ego-vehicle and generate a sequence of control strategy (a reference speed profile in the future) to achieve fuel efficient operation goal. PCC algorithm can be formulated as an optimization problem, which is to find the optimal longitudinal distance trajectory $s^*(t)$ that minimizes:
\begin{equation}
J = \int_0^T \left(\dot{m}_f + k_a \ddot{s}(t)^2 \right)dt
\end{equation}
Subject to:
\begin{equation}
\begin{split}
s_1 = s(0)\\
s_T = s_1 + v_{ref}T \\
\dot{s_1} = \dot{s(0)}\\
\dot{s_T} = v_{ref} \\
\dot{s_i} \leq V_{max}, i\in[2, 3, ..., n-1] \\
s_i \leq s^{fence}_i, i\in[2, 3, ..., n - 1]
\end{split}
\end{equation}
\begin{figure}[htbp] 
	\centering
	\mbox{} \hfill
	\includegraphics[width=1\linewidth]{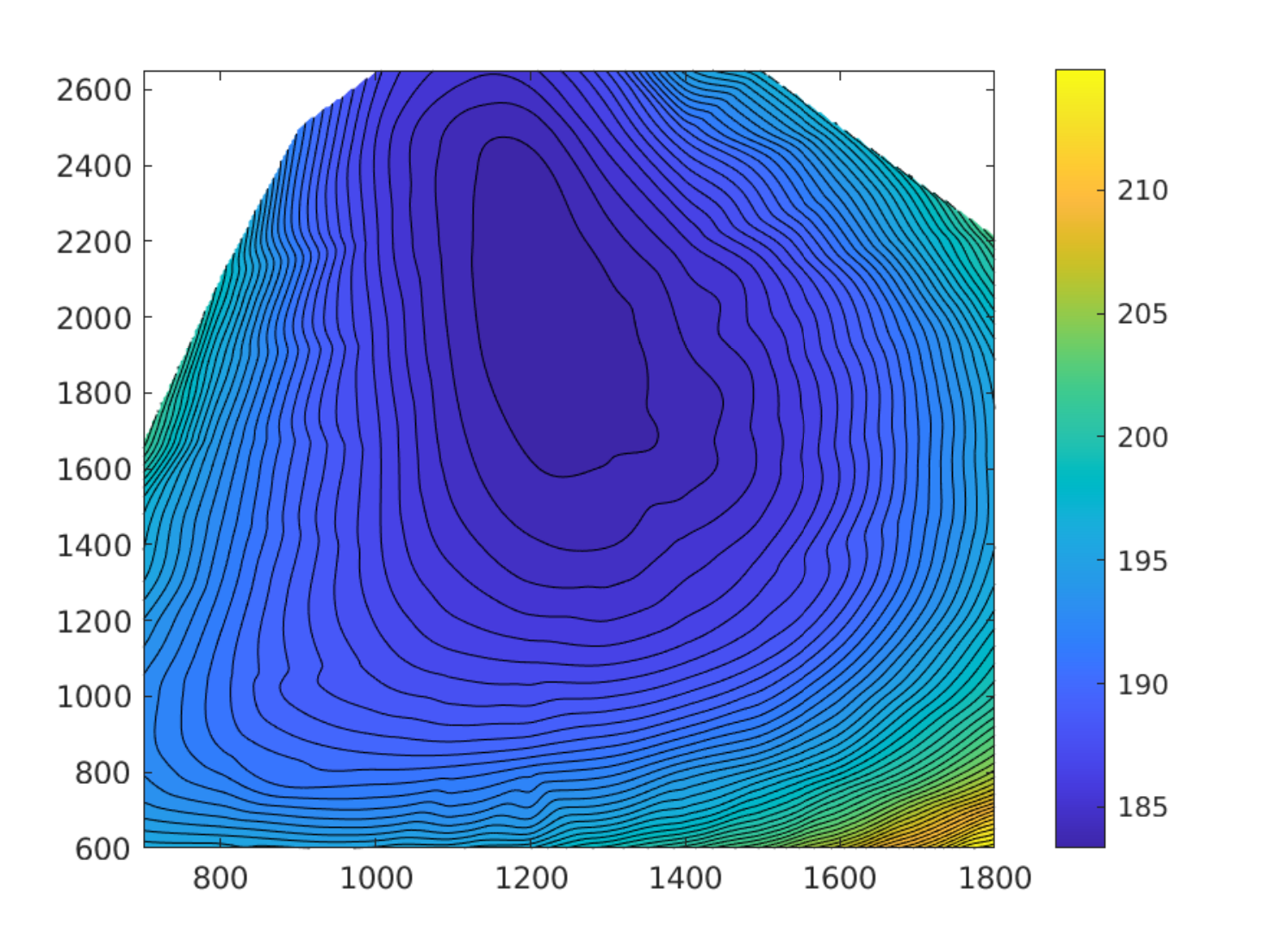}
	\hfill \mbox{}
	\caption{\label{fig:bsfc_map}%
		The diagram of the engine model. The abscissa axis represents the engine speed (rpm), the vertical axis is the engine torque (Nm) and the value (contour) denotes the BSFC (g/kWh).
	}
\end{figure} 
where $\dot{m}_f$ is a nonlinear lookup table that computes the mass fuel flow rate based on the engine power shown in Fig.~\ref{fig:bsfc_map}, $K_a$ is a penalizing factor on acceleration so that the convex nature of $J$ can be ensured. The engine power can be obtained from a vehicle dynamics model, as:
\begin{equation}
    p_e=\dot{s}\eta{(sin(\theta(s))+\mu(\dot{s}))gM_{veh}+\frac{1}{2}\rho_{air} A_f C_d \dot{s}^2+\ddot{s}M_{veh}\dot{s}}
\end{equation}
in which $\eta$ is total power efficiency from engine torque to propulsion force, $\theta(s)$ is road gradient with respect to the distance ahead of vehicle, $\mu(\dot{s})$ is tire rolling friction, $M_{veh}$ indicates the vehicle mass, $\rho_{air}$ is the density of air, $A_f$ means the front area of vehicle and $C_d$ is the air drag constant.
We assume that $\dot{m}_f$ and $\frac{d\dot{m}_f}{dp_e}$ are smooth for $p_e \in [0, p_{e,max}]$, the necessary condition on optimality states that, for any arbitrary small perturbation $\delta s(t)$:
\begin{equation}
\begin{split}
    \int_0^T \left(
\dot{m}_f\left(p_e(s, \dot{s}, \ddot{s})\right) + k_a \ddot{s}^2 
\right)dt 
= \\
\int_0^T \left(
\dot{m}_f\left(p_e(s + \delta s, \dot{s} + \delta \dot{s}, \ddot{s} + \delta \ddot s)) + 
k_a (\ddot{s} + \delta \ddot{s}\right)^2 
\right)dt
\end{split}
\end{equation}
Using Taylor expansion, we have:
\begin{equation}
\begin{split}
    \dot{m}_f(s + \delta s, \dot{s} + \delta \dot{s}, \ddot{s} + \delta \ddot s) = \\
    \dot{m}_f(s, \dot{s}, \ddot{s}) + 
\frac{\partial \dot{m}_f} {\partial s} \delta s +
\frac{\partial \dot{m}_f} {\partial \dot{s}} \frac{d \delta s}{dt} +
\frac{\partial \dot{m}_f} {\partial \ddot{s}} \frac{d^2 \delta s}{dt^2}  + H.O.T.
\end{split}
\end{equation}
Therefore, the necessary condition then takes the following form as:
\begin{equation}
    \int_0^T \left(
\frac{\partial \dot{m}_f} {\partial s} \delta s +
\frac{\partial \dot{m}_f} {\partial \dot{s}} \frac{d \delta s}{dt} +
\frac{\partial \dot{m}_f} {\partial \ddot{s}} \frac{d^2 \delta s}{dt^2} + 
2k_a \frac{d^2 s}{dt^2} \frac{d^2 \delta s}{dt^2}
\right)dt = 0
\end{equation}
Then we employ the well-established finite element method to solve such an optimization problem~\cite{liao2018regularized}. Finally we will get the solution:
\begin{equation}
    X=[s_1, \dot{s}_1, s_2, \dot{s}_2, ..., s_n, \dot{s}_n]^T
\end{equation}
this speed profile is fuel optimal control strategy comparing to constant speed cruise control with same average speed setup. Since solving optimization problem is time consuming, which may cost about 60 second from initialization status to 95\% convergence, meanwhile the optimization significantly slows down as the predictive horizon increases. Hence, we choose to execute PCC algorithm in advance to plan a global optimal speed profile, then the trajectory planner will query the speed profile according to the given coordinates. By doing so, we can not only reduce the operation cost but also accelerate the online planning.

\subsubsection{Trajectory Planner}
\begin{figure}
    \centering
    \begin{subfigure}[b]{1\linewidth}
        \centering
        \includegraphics[width=\linewidth]{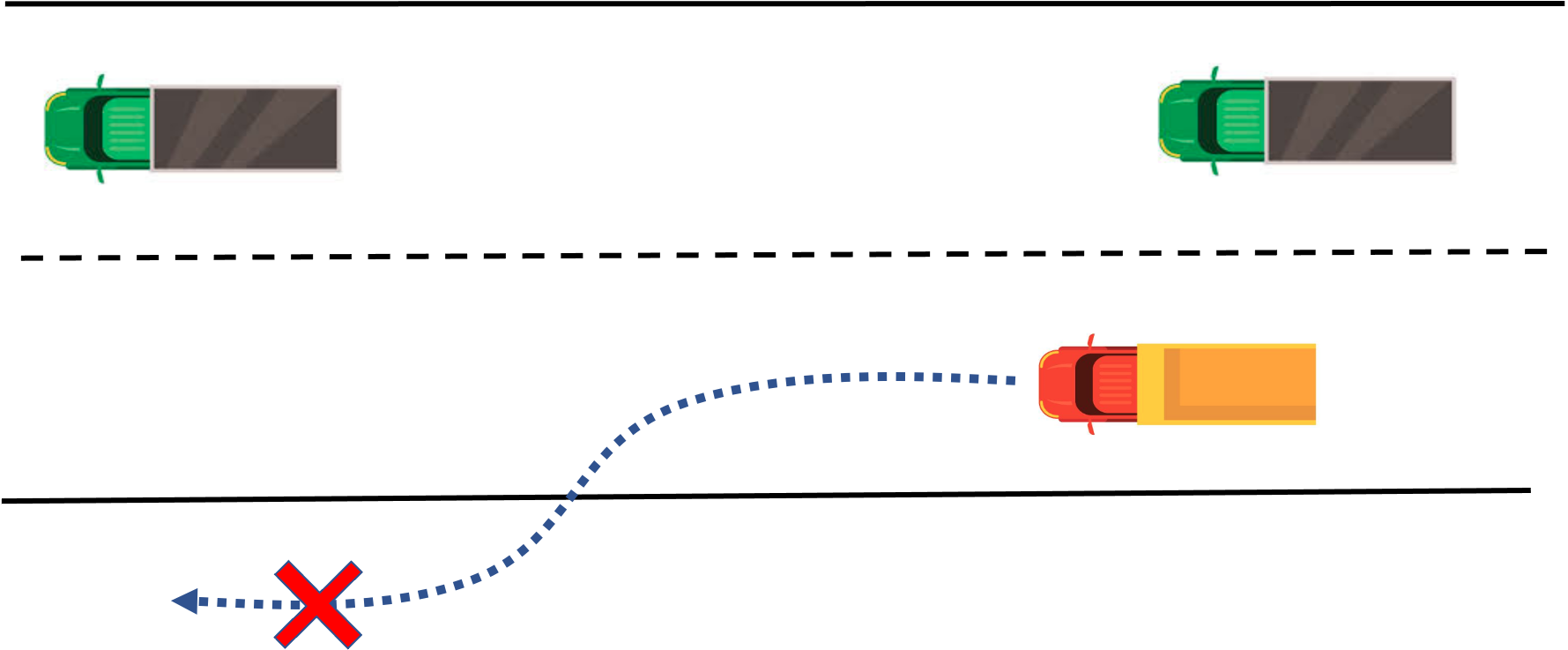}
        \caption[Network2]%
        {{\small Invalid Lane}}    
    \end{subfigure}
    \vskip\baselineskip
    \begin{subfigure}[b]{1\linewidth}  
        \centering 
        \includegraphics[width=\linewidth]{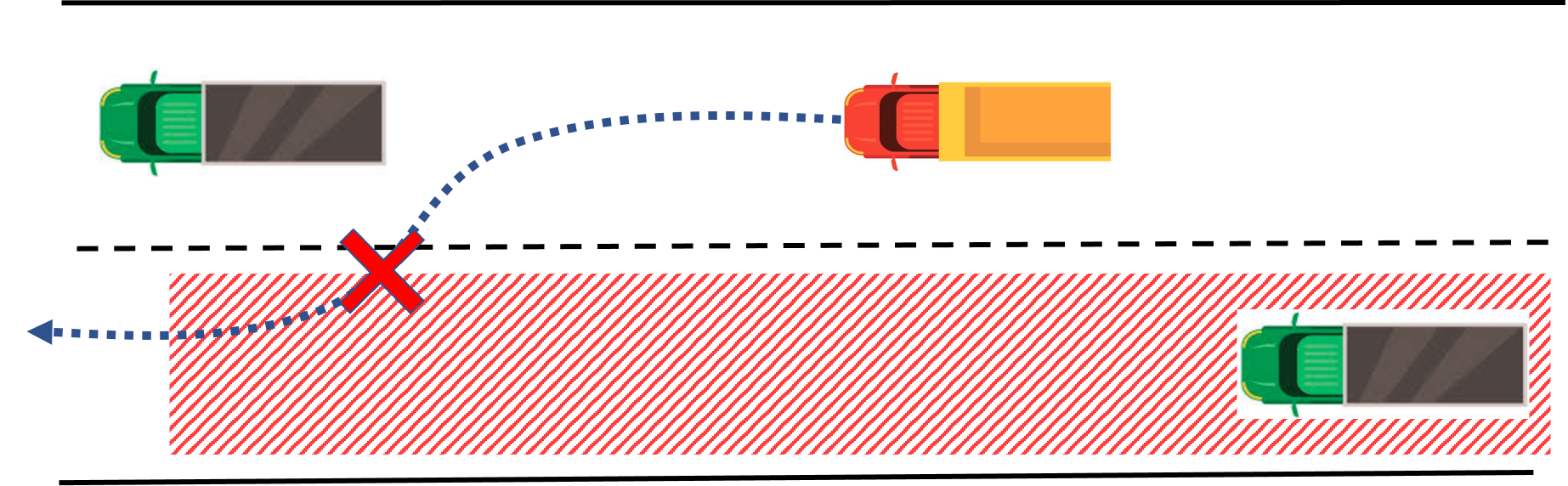}
        \caption[]%
        {{\small Collision Risk Lane Changing}}    
    \end{subfigure}
    \vskip\baselineskip
    \begin{subfigure}[b]{1\linewidth}   
        \centering 
        \includegraphics[width=\linewidth]{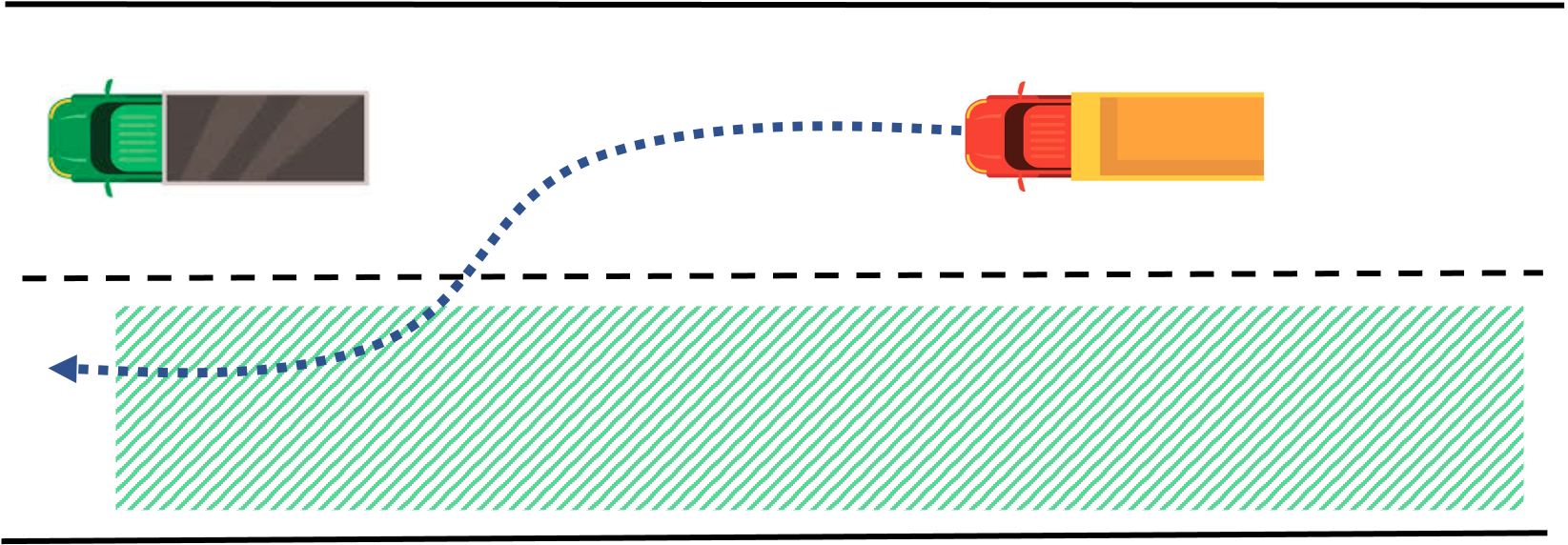}
        \caption[]%
        {{\small Valid and Safe Lane Changing}}
    \end{subfigure}
    \caption
    {\small Three different situations for lane change mode: (a) The target lane is not exist. (b) There is collision risk for lane changing. (c) Target Lane exist and safe for lane changing now.}
    \label{fig:lc}%
\end{figure}

We have two different mode for trajectory planning module. If there is no lane change order sent from the superior module, the lane keeping mode will be activated. Otherwise, if a lane change signal is received, the system will enter the lane change mode. In lane keeping mode, the planner queries the map for the current lane's reference center-line, then discretizes the center-line into waypoints $(s_1, s_2, s_3, ...., s_n)$, then assign the speed profile for each waypoint. In lane change mode, the planner queries the target lane in the map and determine the safety, as shown in Fig.~\ref{fig:lc}, there are three different situations: (a) The target lane is not exist. (b) There is collision risk for lane changing. (c) The target lane exist and safe for lane changing now. If it is invalid or not safe for lane changing, the algorithm will return to the lane keeping mode. If it is suitable for lane changing, we adopt quintic polynomial trajectory planning algorithm~\cite{piazzi2002quintic} to generate a smooth trajectory from the current position to the target waypoint in target lane. Then the planner queries the reference speed profile and assign to the trajectory. The trajectory planner module is summarized in Algo.~\ref{alg:planner}.

\begin{algorithm}[h!]
\caption{Trajectory Planner and Speed Profile Assignment}
\label{alg:planner}
\begin{algorithmic}[1]
\State Receive the high level decision $a_t$
\If{ $a_t = 0$ }
    \State Query for the waypoints from current lane: $(s_1, s_2, s_3, ...., s_n)$
    \State $s_{Collision} = s_{FrontCar} - s_{Ego}$ 
    \For {waypoint $s_i = s_1, s_2, s_3,.....$}
        \If{$s_{Collision}> s_n$}
            \State $v_i$ = Query\_Reference\_Speed$(s_i)$
        \Else
            \State $v_i = Brake$
        \EndIf{}    
    \EndFor
\Else
    \State Check the availability of target lane ($a_t = -1$ is the left lane, $a_t = 1$ is the right lane)
    \If{Target Lane is not available}
        \State Return to Lane Keeping Mode
    \Else
        \State Get the end pose of lane changing $s_{end}$
        \State $(s_1, s_2, s_3, ...., s_n)$ = QuinticPlanner($s_{ego}, s_{end}$)
        \For {waypoint $s_i = s_1, s_2, s_3,.....$}
            \State $v_i$ = Query\_Reference\_Speed$(s_i)$
        \EndFor{}
    \EndIf{}
\EndIf{}
\State Return $(s_1, v_1, s_2, v_2, s_3, v_3,....s_n, v_n)$
\end{algorithmic}
\end{algorithm}
Finally we obtain a sequence of waypoints along with the corresponding speed profile, $(s_1, v_1, s_2, v_2, s_3, v_3,....s_n, v_n)$, which will be sent to next module.

\subsection{Experiments of Intelligent Decision Making}
\subsubsection{Numerical Experiment Scenarios}
\begin{figure}[htbp] 
\centering
    \begin{subfigure}[b]{0.49\linewidth}
        \centering
        \includegraphics[width=\linewidth]{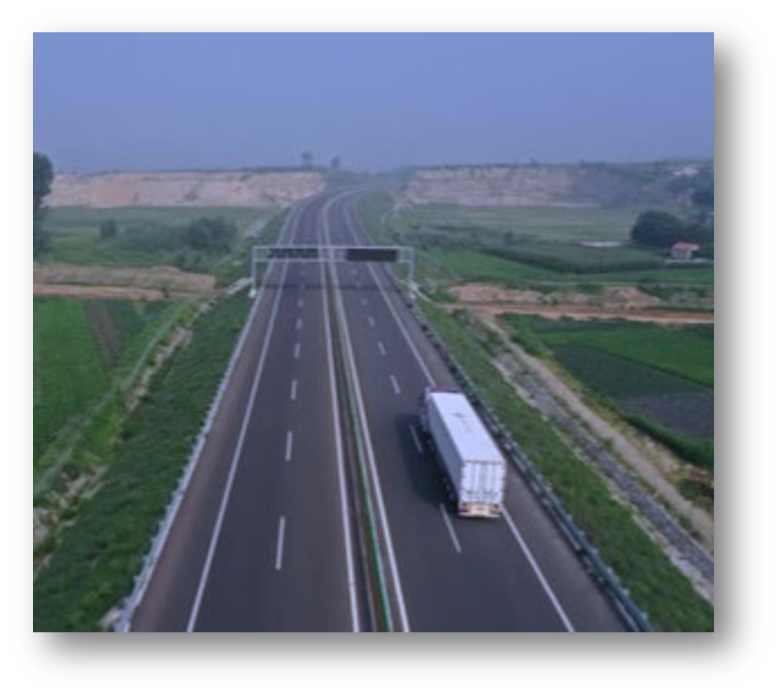}
        \caption[Network2]%
        {{\small Jinan Test Site}}    
    \end{subfigure}
    \hfill
    \begin{subfigure}[b]{0.49\linewidth}  
        \centering 
        \includegraphics[width=\linewidth]{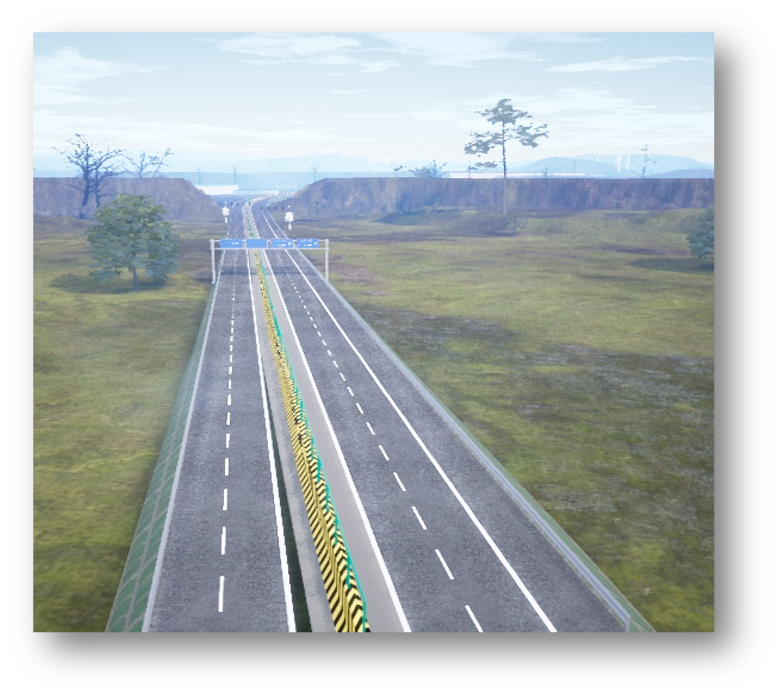}
        \caption[]%
        {{\small Reconstructed Test Scenario}}    
    \end{subfigure}
	\caption{\label{fig:jinan_virtual_real}%
		We reconstruct our test site (a closed highway in Jinan Eastern China) in our system for numerical experiment. 
	}
\end{figure}

\begin{figure}
    \centering
    \begin{subfigure}[b]{0.475\linewidth}
        \centering
        \includegraphics[width=\linewidth]{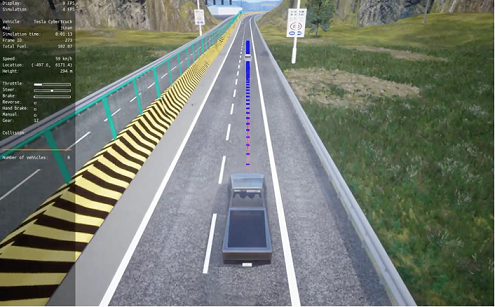}
        \caption[Network2]%
        {{\small Car Following}}    
    \end{subfigure}
    \hfill
    \begin{subfigure}[b]{0.475\linewidth}  
        \centering 
        \includegraphics[width=\linewidth]{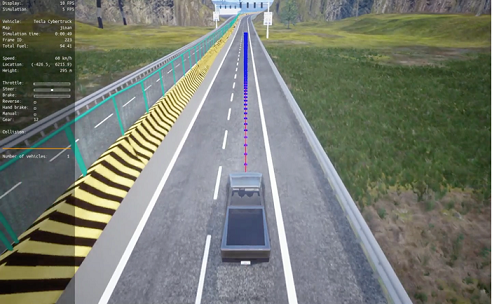}
        \caption[]%
        {{\small Running alone}}    
    \end{subfigure}
    \vskip\baselineskip
    \begin{subfigure}[b]{0.475\linewidth}   
        \centering 
        \includegraphics[width=\linewidth]{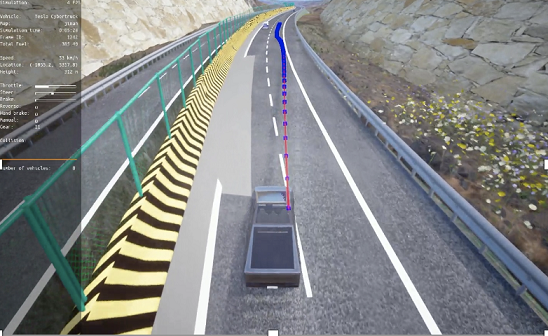}
        \caption[]%
        {{\small Lane Change}}    
    \end{subfigure}
    \hfill
    \begin{subfigure}[b]{0.475\linewidth}   
        \centering 
        \includegraphics[width=\linewidth]{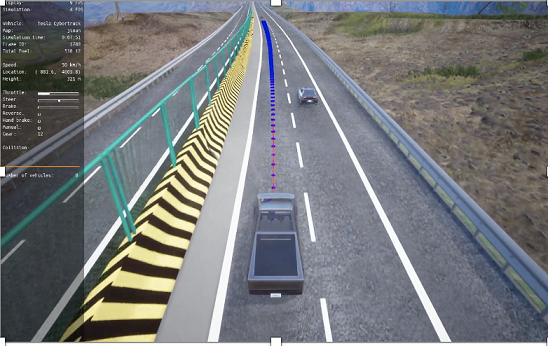}
        \caption[]%
        {{\small Overtaking}}    
    \end{subfigure}
    \caption
    {\small We rebuild the texture in visualization, illustrate our ego-truck, traffic flow and also the trajectory on the screen.}
    \label{fig:carla}%
\end{figure}

We reconstruct our test site, which is a closed highway in Jinan, a city in Eastern China, shown in Fig.~\ref{fig:jinan_virtual_real} and also rebuild the texture in visualization. We illustrate the ego-truck, the traffic flow (of neighbor cars) and the trajectory on the screen, as shown in Fig.~\ref{fig:carla}. The test site is 15km length in total, covering variable typical road condition, e.g., slope, tunnel, curve.

\subsubsection{Experiment Setup}
\begin{figure}[htbp] 
	\centering
	\mbox{} \hfill
	\includegraphics[ width=1\linewidth]{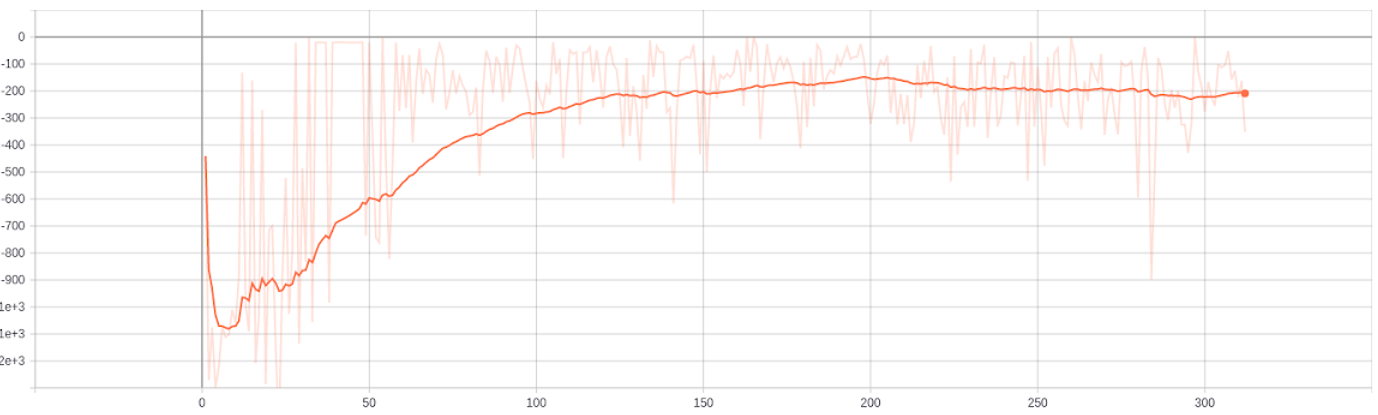}
	\hfill \mbox{}
	\caption{\label{fig:rewardcurve}%
		Average accumulative reward for each epoch during the training procedure. The vertical axis represents the average accumulative reward of each epoch, the horizontal axis represents number of epoch.
	}
\end{figure}

We build our reinforcement learning based decision making model based on Pytorch~\cite{paszke2019pytorch} and train it on a ThinkStation P920 with Intel Xeon(R) Silver 4110 2.1GHz x32 and NVIDIA RTX 2080Ti. During the training process, the learning rate is fixed as $5\mathrm{e}{-4}$, the optimizer is ADAM~\cite{kingma2014adam}, and the training batch size sets as $64$. The proposed system is running at \SI{10}{Hz}. Note that up to six cars in front of the ego-truck and two cars behind the ego-truck are observable. The model is trained for $2$ days, and the reward curve is shown in~\prettyref{fig:rewardcurve}.

For performance evaluation of our RL model, we present the following metrics for quantitative evaluation:
\begin{enumerate}
\item Delta Velocity: the average difference between ego-truck speed $V$ and reference speed $V_{ref}$. It is formulated as: $\nabla V = (V_{ref} - V)/V_{ref}$.
\item Count of Lane Change: the average count of lane change decision during the journey.
\end{enumerate}

To increase the diversity of the environment, we investigate traffic with three different densities and two average speeds. We modify the traffic spawn probability as $0.05$, $0.02$, $0.005$, corresponding to the dense, medium, sparse density respectively. For the average speed of traffic aspect, we set the maximum speed of the traffic flow as \SI{12.5}{m/s}, \SI{15}{m/s}, and our ego-truck's reference speed is fixed as \SI{16.67}{m/s} (\SI{60}{km/h}) which is suitable for most logistics operations. 

For comparison, we propose three baseline algorithms with rule-based FSM to mimic different characteristic drivers' behavior. When the ego-truck detects obstacles $d$\si{m} ahead of it, we will check whether there is no traffic in the neighbor lane with $d$\ \si{m} ahead of it and $d$\si{m} back of it. If so, the lane change decision will be sent to the trajectory planner, if not, we will keep in this lane. Here, we use $d$ to describe the drivers' characteristics: aggressive, neutral and conservative, corresponding to $d=50, 100$ and $150$.

\begin{table*}[!h]
\centering
\begin{tabular}{@{}|l|c|l|l|l|l|l|l|l|l|@{}}
\toprule
\multicolumn{1}{|c|}{\multirow{2}{*}{Scenario}} &
  \multicolumn{1}{l|}{\multirow{2}{*}{\begin{tabular}[c]{@{}l@{}}Trafic \\ Speed\end{tabular}}} &
  \multicolumn{4}{c|}{Average Delta Speed} &
  \multicolumn{4}{c|}{Average Count of Lane Change} \\ \cmidrule(l){3-10} 
\multicolumn{1}{|c|}{} &
  \multicolumn{1}{l|}{} &
  Conservative &
  Neutral &
  Aggressive &
  RL-Baseline &
  Conservative &
  Neutral &
  Aggressive &
  RL-Baseline \\ \midrule
\multirow{3}{*}{Dense}
                        & 12.5m/s  &  38.9\%/0.059 &  28.5\%/0.043  & 24.5\%/0.038 
& \textbf{24.0\%}/0.037  &  2.84/1.76   &  6.23/2.67    & 6.98/3.87  & {6.16}/2.65  \\ \cmidrule(l){2-10} 
                        & 15m/s   &  30.9\%/0.047 &  20.1\%/0.031  & 18.8\%/0.028 
& \textbf{17.3\%}/0.027   &  3.06/1.82   &  6.89/2.85    &  7.13/3.91   & {6.24}/2.68 \\ \midrule
\multirow{3}{*}{Medium} 
                        & 12.5m/s &  14.1\%/0.022   &   13.8\%/0.021 & {6.6\%}/0.010  
& 8.0\%/0.012  &  4.58/1.23    & 5.09/2.37    &  8.93/3.39    & {6.22}/2.67  \\ \cmidrule(l){2-10} 
                        & 15m/s   &  4.4\%/0.007  &  1.6\%/0.002  &  1.6\%/0.002  
& \textbf{1.4\%}/0.002  &  5.00/1.34  & 4.02/2.08   &  8.99/3.41    & {6.30}/2.69  \\ \midrule
\multirow{3}{*}{Sparse}  
                        & 12.5m/s &  5.9\%/0.009 &  3.4\%/0.005   & 4.4\%/0.007   
& \textbf{3.4\%}/0.005  &  2.55/0.68    & 2.83/0.75   &  2.99/0.80    & {3.02}/0.81  \\ \cmidrule(l){2-10} 
                        & 15m/s   & 1.0\%/0.001  & 1.0\%/0.001   &  10.0\%/0.002  
& \textbf{1.0\%}/0.001  & 0.40/0.11    & 0.40/0.10    & 0.50/0.13    & 0.77/0.21   \\ \bottomrule
\end{tabular}
\caption{Average Delta Velocity (shown as mean/std) evaluated for different methods on different scenarios, and the best results in each category are in \textbf{bold}. Note that Conservative in table is the abbreviation of the conservative baseline. Average Count of Lane Change (shown as mean/std) evaluated for different methods on different scenarios.} 
\label{tab:lccount}
\end{table*}

We compare our RL based lane change decision-maker with aggressive, neutral and conservative baseline algorithms. Note that except the decision-maker module, all other modules remain unchanged. We run each experiment trial $30$ times and compute the mean value of them.

Tab.~\ref{tab:lccount} shows the performance evaluated using different methods in different test scenarios. It can be seen that our RL base method yields the best performance according to delta velocity metrics in most cases. We noticed that our method's performance is worse than the aggressive baseline algorithm. This is because that aggressive baseline tends to change lane frequently. Ideally, if the ego-truck can change the lane frequently enough that the influence of any car in front of it can be avoided. However, because of the character of the truck we mentioned in the introduction section, it is impossible, even dangerous, for a truck to be tap-dancing through the traffic flow. That is the reason why we add a penalty for all lane change decisions in our reward function for the RL. Our goal is to achieve the reference speed with as few lane changes as possible while ensuring safety. We can observe that our RL based approach changes lane fewer than the aggressive baseline algorithm in most cases. With fewer change, we also achieve comparable performance compared to baseline methods. 

\subsection{Experiment of Fuel Efficiency}
\begin{figure*}[htbp] 
	\centering
	\mbox{} \hfill
	\includegraphics[width=1\linewidth]{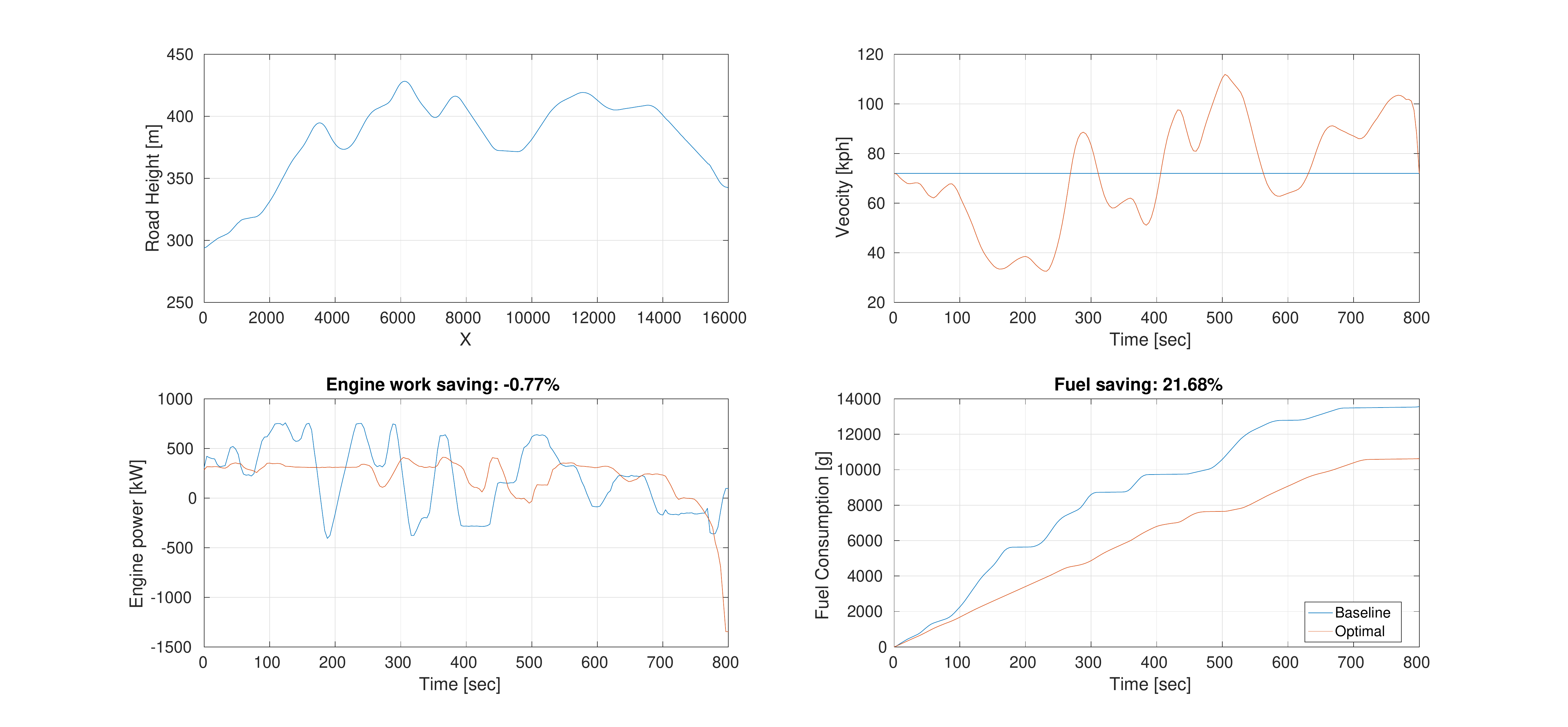}
	\hfill \mbox{}
	\caption{\label{fig:pcc}%
		Fuel saving experiment in static environment. Top left: the road slope (m) of our test site. Top right: the velocity (Km/h). Bottom left: the engine power (kW) during the test. Bottom right: the accumulative fuel consumption (g).
	}
\end{figure*}

We evaluate the fuel efficient speed planner described in Sec.~\ref{sec:fuel_saving}. First, we conduct the experiment in static environment, in which the traffic is not involved. The baseline method we choose to compare with is constant speed cruise strategy (72 km/h). As shown in Fig.~\ref{fig:pcc}, the top left figure illustrates the road slope (m) of our test site, the top right figure shows the velocity (Km/h) of our proposed system (orange line) and baseline method (blue line), the bottom left figure shows the engine power (kW) during the test, the bottom right figure illustrates the accumulative fuel consumption (g) for our method and baseline method. The results show that our proposed system can effectively allocate the engine working status, which is more stable than baseline method. The key fuel saving capacity for heavy duty truck on hilly road is to utilize the conversion between kinetic energy and potential energy in uphill and downhill. The velocity allocation of our proposed system is shown in top right figure, which decelerate in uphill process and accelerate in downhill process. Our proposed system can save about 21.68\% fuel comparing to constant speed cruise strategy. 
 
\section{Real World Experiment}
\label{sec:real_truck_exp}
We deploy our proposed intelligent autonomous truck system to the real truck, and conduct several experiments in our test site in Jinan. In this section, we describe the setup detail of the real truck deployment firstly, then we illustrate several results of the real world experiment.

In order to integrate our proposed system to real truck, we plug our system into Inceptio autonomous driving platform to obtain the perception result and control the real truck, as shown in Fig.~\ref{fig:deploy} 

\begin{figure}
    \centering
    \begin{subfigure}[b]{0.475\linewidth}
        \centering
        \includegraphics[width=\linewidth]{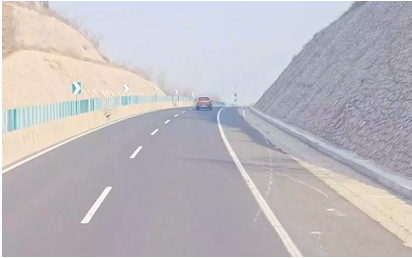}
        \caption[]%
        {{\small Real Truck Deployment}}    
    \end{subfigure}
    \hfill
    \begin{subfigure}[b]{0.475\linewidth}  
        \centering 
        \includegraphics[width=\linewidth]{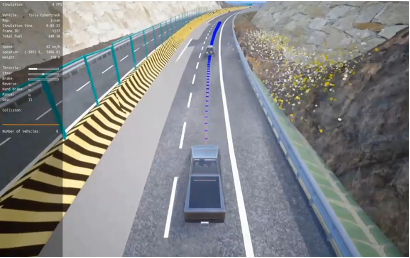}
        \caption[]%
        {{\small Simulation Visualization}}    
    \end{subfigure}
    \vskip\baselineskip
    \begin{subfigure}[b]{0.475\linewidth}   
        \centering 
        \includegraphics[width=\linewidth]{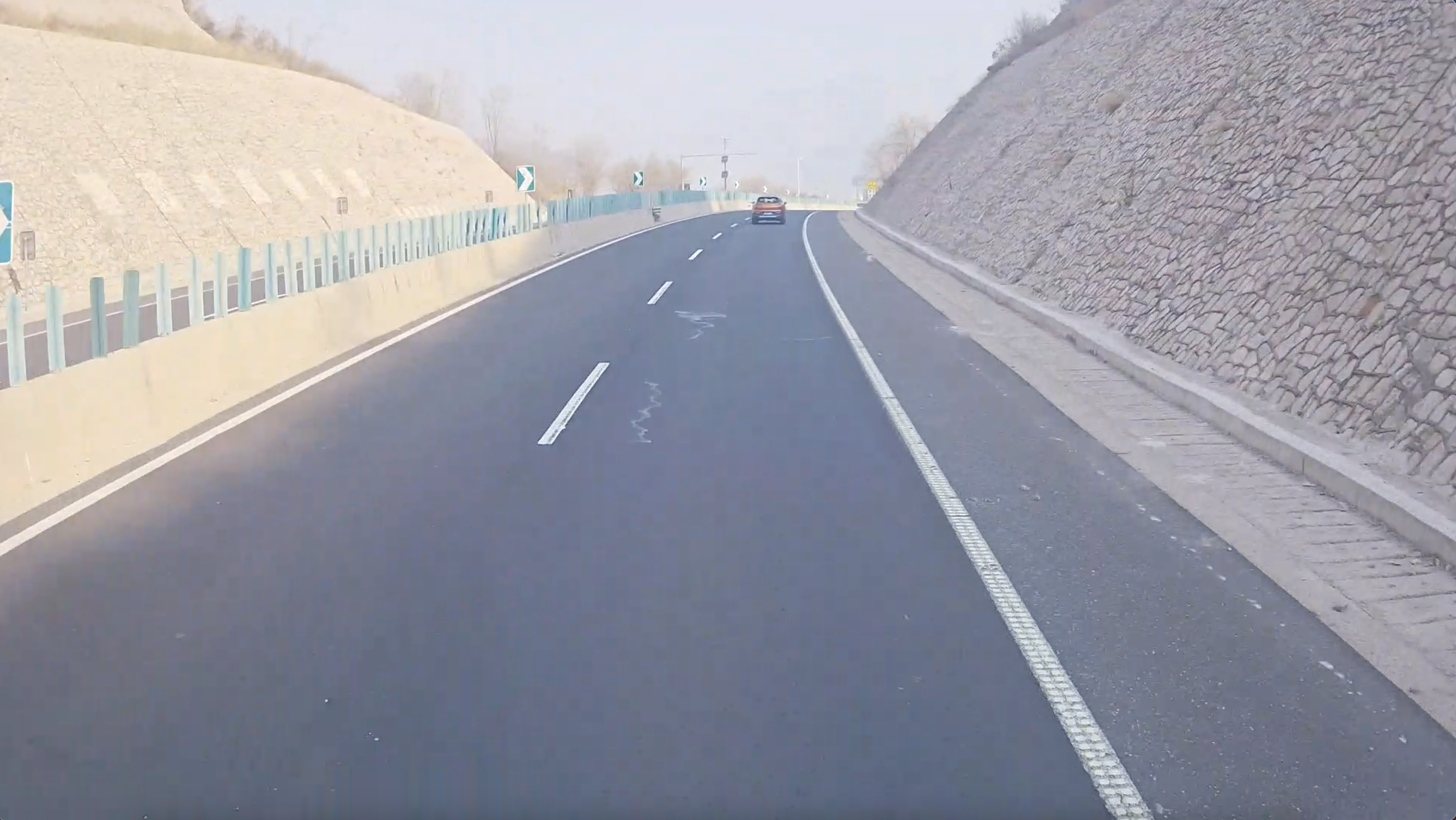}
        \caption[]%
        {{\small Car Following (Real)}}    
    \end{subfigure}
    \hfill
    \begin{subfigure}[b]{0.475\linewidth}   
        \centering 
        \includegraphics[width=\linewidth]{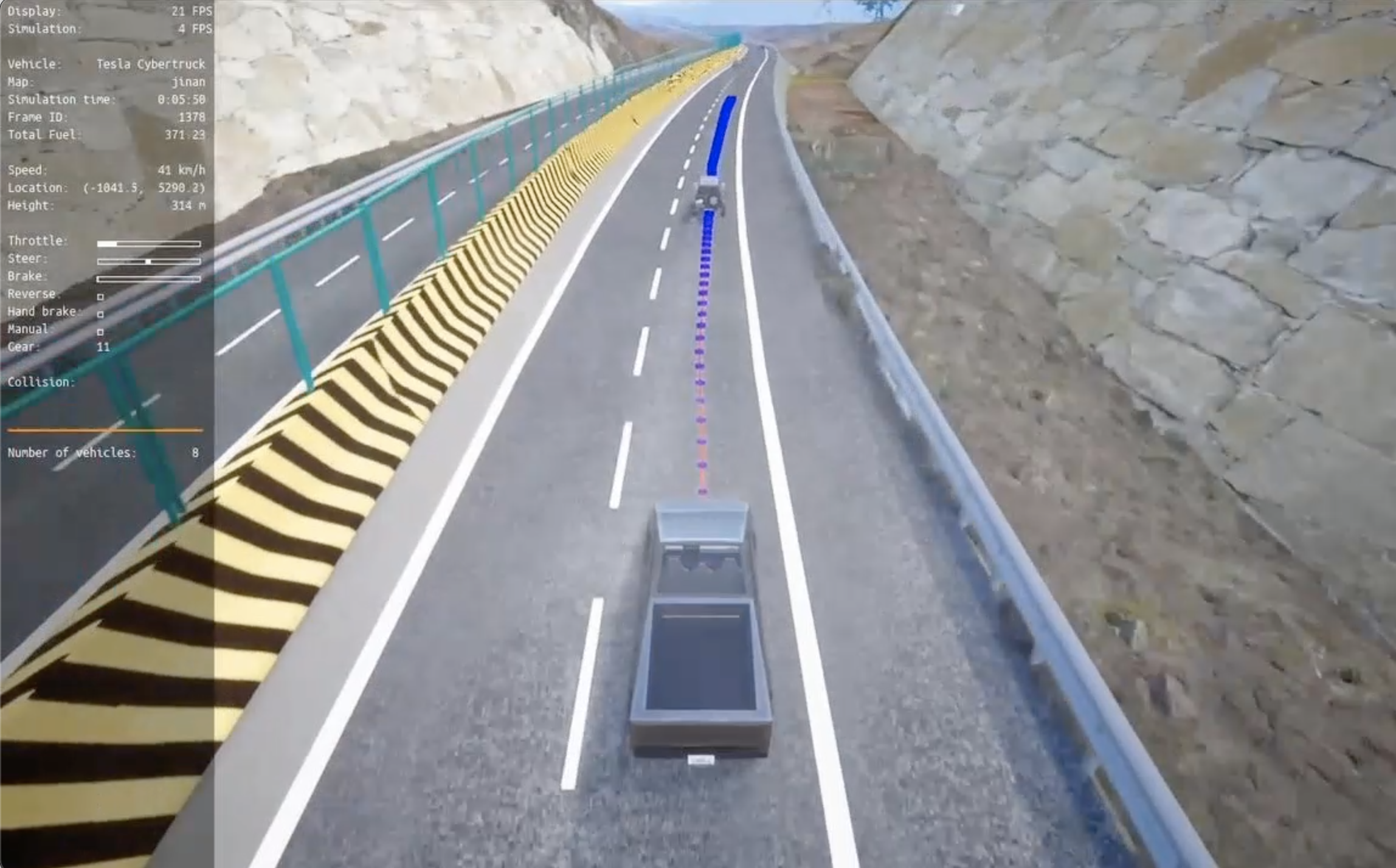}
        \caption[]%
        {{\small Car Following (Sim)}}    
    \end{subfigure}
        \vskip\baselineskip
    \begin{subfigure}[b]{0.475\linewidth}   
        \centering 
        \includegraphics[width=\linewidth]{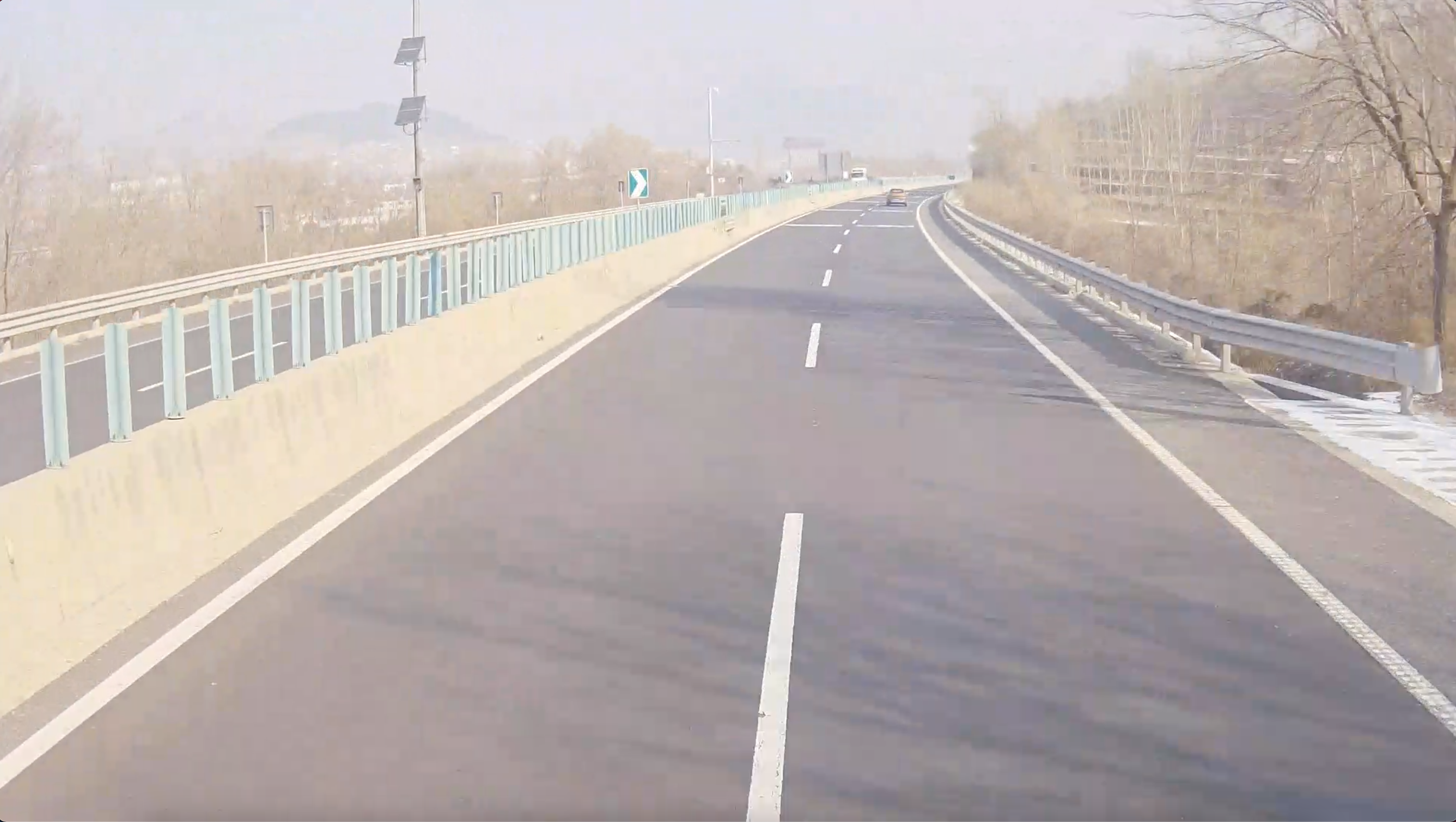}
        \caption[]%
        {{\small Lane Changing (Real)}}    
    \end{subfigure}
    \hfill
    \begin{subfigure}[b]{0.475\linewidth}   
        \centering 
        \includegraphics[width=\linewidth]{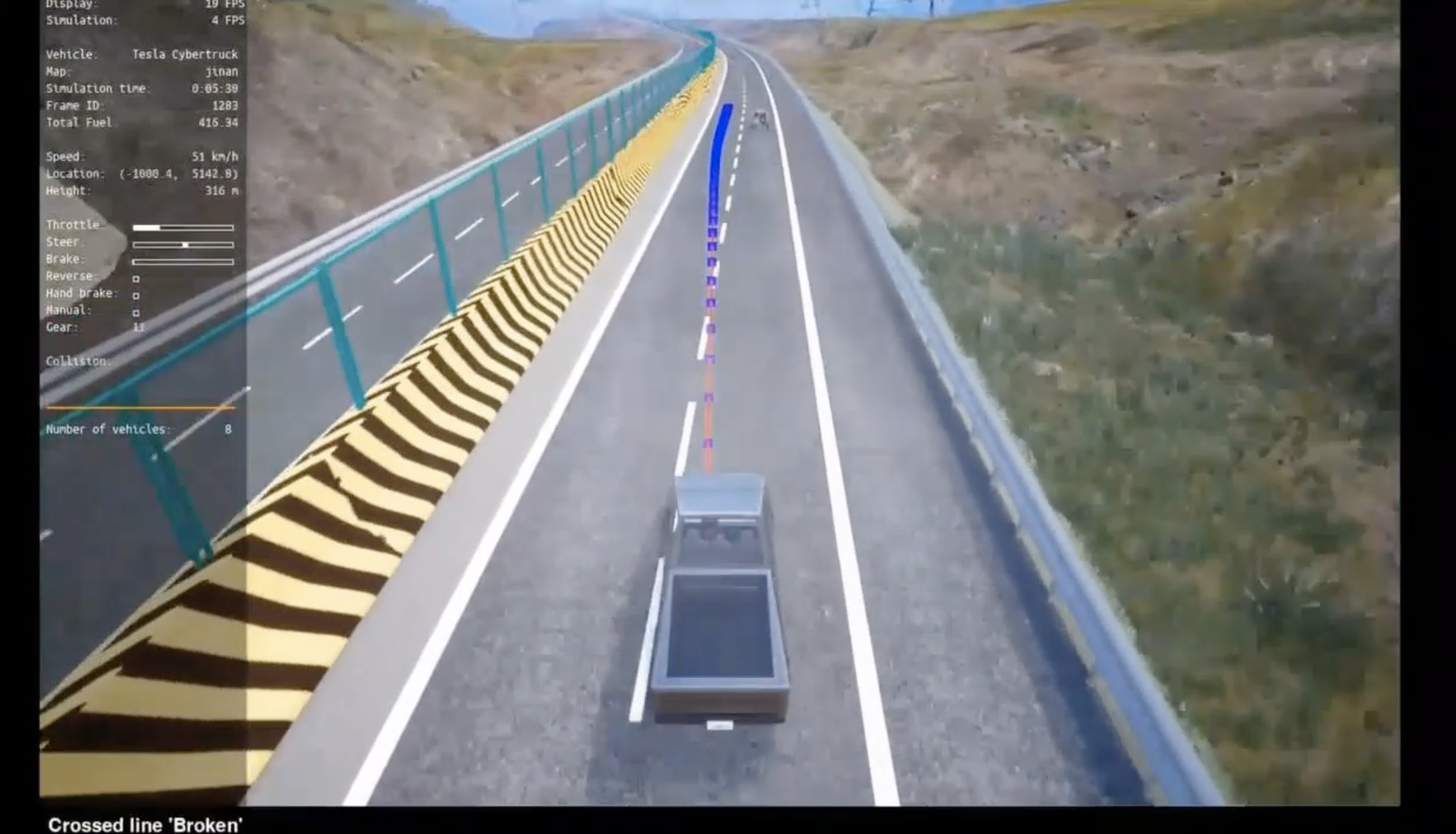}
        \caption[]%
        {{\small Lane Changing (Sim)}}    
    \end{subfigure}
    \caption
    {\small We deploy our intelligent self-driving truck system to real truck, and conduct a real world experiment at our test site in Jinan. (Left) Jinan test site is a closed highway, we arrange some actor cars on the road during our testing. We also use our visualizer to monitor our system's operating status. (Right)}
    \label{fig:sim_real}%
\end{figure}

We recorded a video from the front camera on our truck, and also the visualization of our system's running. As shown in Fig.~\ref{fig:sim_real}, our system demonstrate robust performance on real truck experiment as good as simulated environment. 

\section{Conclusions and Future Work}
In this paper, we describe an intelligent self-driving truck system, which
consists of three main components: the realistic traffic simulation module for generating realistic traffic flow in testing scenarios, the high-fidelity truck model for mimicking real truck response in real world deployment and the intelligent planning module with learning-based decision making algorithm and multi-mode trajectory planner. We conduct adequate evaluation experiments for each components and the results show the robust performance of our proposed intelligent self-driving truck system. Our system is the first open-sourced full self-driving truck system for logistic operation and mass production. In addition, the high-fidelity truck model filled the gap and demand for self-driving truck development in industry and academia. Our code is available at https://github.com/InceptioResearch/IITS

In this paper, we only focus on self-driving truck in highway transportation scenarios, covering 90\% logistic daily operation. However, there is still a long way to full L4 autonomous driving truck, especially dealing with more complex traffic and pedestrians. We noticed many recent advanced works have been released targeting to several areas in autonomous driving, e.g., collision avoidance~\cite{zhang2021safe}, left-turn planning~\cite{shu2020autonomous}. For the future work, we plan to cover more logistic truck operation scenarios in addition to highway, for example on-ramp/off-ramp scenarios, left-turn/right-turn planning in crowd intersection. We also welcome you to contribute codes and ideas to our project.

{\small
\bibliographystyle{IEEEtran}
 \typeout{}
\bibliography{references}

\begin{thebibliography}{10}
\providecommand{\url}[1]{#1}
\csname url@samestyle\endcsname
\providecommand{\newblock}{\relax}
\providecommand{\bibinfo}[2]{#2}
\providecommand{\BIBentrySTDinterwordspacing}{\spaceskip=0pt\relax}
\providecommand{\BIBentryALTinterwordstretchfactor}{4}
\providecommand{\BIBentryALTinterwordspacing}{\spaceskip=\fontdimen2\font plus
\BIBentryALTinterwordstretchfactor\fontdimen3\font minus
  \fontdimen4\font\relax}
\providecommand{\BIBforeignlanguage}[2]{{%
\expandafter\ifx\csname l@#1\endcsname\relax
\typeout{** WARNING: IEEEtran.bst: No hyphenation pattern has been}%
\typeout{** loaded for the language `#1'. Using the pattern for}%
\typeout{** the default language instead.}%
\else
\language=\csname l@#1\endcsname
\fi
#2}}
\providecommand{\BIBdecl}{\relax}
\BIBdecl

\bibitem{viscelli2018driverless}
S.~Viscelli, ``Driverless? autonomous trucks and the future of the american
  trucker,'' 2018.

\bibitem{fortune2020market}
F.~B. Insights, ``Autonomous trucks market,'' 2020.

\bibitem{ulbrich2013probabilistic}
S.~Ulbrich and M.~Maurer, ``Probabilistic online pomdp decision making for lane
  changes in fully automated driving,'' in \emph{16th International IEEE
  Conference on Intelligent Transportation Systems (ITSC 2013)}.\hskip 1em plus
  0.5em minus 0.4em\relax IEEE, 2013, pp. 2063--2067.

\bibitem{wang2019lane}
J.~Wang, Q.~Zhang, D.~Zhao, and Y.~Chen, ``Lane change decision-making through
  deep reinforcement learning with rule-based constraints,'' in \emph{2019
  International Joint Conference on Neural Networks (IJCNN)}.\hskip 1em plus
  0.5em minus 0.4em\relax IEEE, 2019, pp. 1--6.

\bibitem{codevilla2018end}
F.~Codevilla, M.~M{\"u}ller, A.~L{\'o}pez, V.~Koltun, and A.~Dosovitskiy,
  ``End-to-end driving via conditional imitation learning,'' in \emph{2018 IEEE
  International Conference on Robotics and Automation (ICRA)}.\hskip 1em plus
  0.5em minus 0.4em\relax IEEE, 2018, pp. 4693--4700.

\bibitem{sallab2017deep}
A.~E. Sallab, M.~Abdou, E.~Perot, and S.~Yogamani, ``Deep reinforcement
  learning framework for autonomous driving,'' \emph{Electronic Imaging}, vol.
  2017, no.~19, pp. 70--76, 2017.

\bibitem{lu2004heavy}
X.~Lu, S.~Shladover, and J.~Hedrick, ``Heavy-duty truck control: Short
  inter-vehicle distance following,'' in \emph{Proceedings of the 2004 American
  Control Conference}, vol.~5.\hskip 1em plus 0.5em minus 0.4em\relax IEEE,
  2004, pp. 4722--4727.

\bibitem{Dosovitskiy17}
A.~Dosovitskiy, G.~Ros, F.~Codevilla, A.~Lopez, and V.~Koltun, ``{CARLA}: {An}
  open urban driving simulator,'' in \emph{Proceedings of the 1st Annual
  Conference on Robot Learning}, 2017, pp. 1--16.

\bibitem{shah2018airsim}
S.~Shah, D.~Dey, C.~Lovett, and A.~Kapoor, ``Airsim: High-fidelity visual and
  physical simulation for autonomous vehicles,'' in \emph{Field and service
  robotics}.\hskip 1em plus 0.5em minus 0.4em\relax Springer, 2018, pp.
  621--635.

\bibitem{nvidia-drive-constellation}
NVIDIA, ``Nvidia drive constellation: Virtual reality autonomous vehicle
  simulator,'' 2017.

\bibitem{waymo-carcraft}
A.~C. Madrigal, ``Inside waymo’s secret world for training self-driving
  cars,''
  \url{www.theatlantic.com/technology/archive/2017/08/inside-waymos-secret-testing-and-simulation-facilities/537648/},
  2017.

\bibitem{li2019aads}
W.~Li, C.~Pan, R.~Zhang, J.~Ren, Y.~Ma, J.~Fang, F.~Yan, Q.~Geng, X.~Huang,
  H.~Gong \emph{et~al.}, ``\text{AADS}: Augmented autonomous driving simulation
  using data-driven algorithms,'' \emph{Science robotics}, vol.~4, no.~28,
  2019.

\bibitem{behrisch2011sumo}
M.~Behrisch, L.~Bieker, J.~Erdmann, and D.~Krajzewicz, ``Sumo--simulation of
  urban mobility: an overview,'' in \emph{Proceedings of SIMUL 2011, The Third
  International Conference on Advances in System Simulation}.\hskip 1em plus
  0.5em minus 0.4em\relax ThinkMind, 2011.

\bibitem{fellendorf2010microscopic}
M.~Fellendorf and P.~Vortisch, ``Microscopic traffic flow simulator vissim,''
  in \emph{Fundamentals of traffic simulation}.\hskip 1em plus 0.5em minus
  0.4em\relax Springer, 2010, pp. 63--93.

\bibitem{highway-env}
E.~Leurent, ``An environment for autonomous driving decision-making,''
  \url{https://github.com/eleurent/highway-env}, 2018.

\bibitem{cai2020summit}
P.~Cai, Y.~Lee, Y.~Luo, and D.~Hsu, ``Summit: A simulator for urban driving in
  massive mixed traffic,'' in \emph{2020 IEEE International Conference on
  Robotics and Automation (ICRA)}.\hskip 1em plus 0.5em minus 0.4em\relax IEEE,
  2020, pp. 4023--4029.

\bibitem{trucksim}
M.~S. Corporation, ``Trucksim - mechanical simulation,''
  \url{https://www.carsim.com/products/trucksim/}, 2021.

\bibitem{euro-truck}
M.~T. .~B. AG, ``Euro truck simulator 2,''
  \url{https://eurotrucksimulator2.com}, 2021.

\bibitem{europilot}
MarsAuto, ``Europilot,'' \url{https://github.com/marsauto/europilot}, 2017.

\bibitem{schwarting2018planning}
W.~Schwarting, J.~Alonso-Mora, and D.~Rus, ``Planning and decision-making for
  autonomous vehicles,'' \emph{Annual Review of Control, Robotics, and
  Autonomous Systems}, vol.~1, pp. 187--210, 2018.

\bibitem{wang2018reinforcement}
P.~Wang, C.-Y. Chan, and A.~de~La~Fortelle, ``A reinforcement learning based
  approach for automated lane change maneuvers,'' in \emph{2018 IEEE
  Intelligent Vehicles Symposium (IV)}.\hskip 1em plus 0.5em minus 0.4em\relax
  IEEE, 2018, pp. 1379--1384.

\bibitem{piazzi2002quintic}
A.~Piazzi, C.~L. Bianco, M.~Bertozzi, A.~Fascioli, and A.~Broggi, ``Quintic
  g/sup 2/-splines for the iterative steering of vision-based autonomous
  vehicles,'' \emph{IEEE Transactions on Intelligent Transportation Systems},
  vol.~3, no.~1, pp. 27--36, 2002.

\bibitem{ferguson2008motion}
D.~Ferguson, T.~M. Howard, and M.~Likhachev, ``Motion planning in urban
  environments,'' \emph{Journal of Field Robotics}, vol.~25, no. 11-12, pp.
  939--960, 2008.

\bibitem{urmson2008autonomous}
C.~Urmson, J.~Anhalt, D.~Bagnell, C.~Baker, R.~Bittner, M.~Clark, J.~Dolan,
  D.~Duggins, T.~Galatali, C.~Geyer \emph{et~al.}, ``Autonomous driving in
  urban environments: Boss and the urban challenge,'' \emph{Journal of Field
  Robotics}, vol.~25, no.~8, pp. 425--466, 2008.

\bibitem{treiber2001microsimulations}
M.~Treiber and D.~Helbing, ``Microsimulations of freeway traffic including
  control measures,'' 2001.

\bibitem{kesting2007general}
A.~Kesting, M.~Treiber, and D.~Helbing, ``General lane-changing model mobil for
  car-following models,'' \emph{Transportation Research Record}, vol. 1999,
  no.~1, pp. 86--94, 2007.

\bibitem{chao2020survey}
Q.~Chao, H.~Bi, W.~Li, T.~Mao, Z.~Wang, M.~C. Lin, and Z.~Deng, ``A survey on
  visual traffic simulation: Models, evaluations, and applications in
  autonomous driving,'' in \emph{Computer Graphics Forum}, vol.~39,
  no.~1.\hskip 1em plus 0.5em minus 0.4em\relax Wiley Online Library, 2020, pp.
  287--308.

\bibitem{sewall2011interactive}
J.~Sewall, D.~Wilkie, and M.~C. Lin, ``Interactive hybrid simulation of
  large-scale traffic,'' in \emph{Proceedings of the 2011 SIGGRAPH Asia
  Conference}, 2011, pp. 1--12.

\bibitem{sae1939recommended}
J.~SAE, ``Recommended practice for a serial control and communications vehicle
  network,'' \emph{SAE J1939 Standards Collection}, 1939.

\bibitem{van2015deep}
H.~Van~Hasselt, A.~Guez, and D.~Silver, ``Deep reinforcement learning with
  double q-learning,'' \emph{arXiv preprint arXiv:1509.06461}, 2015.

\bibitem{hasselt2010double}
H.~V. Hasselt, ``Double q-learning,'' in \emph{Advances in neural information
  processing systems}, 2010, pp. 2613--2621.

\bibitem{mnih2015human}
V.~Mnih, K.~Kavukcuoglu, D.~Silver, A.~A. Rusu, J.~Veness, M.~G. Bellemare,
  A.~Graves, M.~Riedmiller, A.~K. Fidjeland, G.~Ostrovski \emph{et~al.},
  ``Human-level control through deep reinforcement learning,'' \emph{nature},
  vol. 518, no. 7540, pp. 529--533, 2015.

\bibitem{lattemann2004predictive}
F.~Lattemann, K.~Neiss, S.~Terwen, and T.~Connolly, ``The predictive cruise
  control--a system to reduce fuel consumption of heavy duty trucks,''
  \emph{SAE transactions}, pp. 139--146, 2004.

\bibitem{liao2018regularized}
D.~Liao-McPherson, M.~Huang, and I.~Kolmanovsky, ``A regularized and smoothed
  fischer--burmeister method for quadratic programming with applications to
  model predictive control,'' \emph{IEEE Transactions on Automatic Control},
  vol.~64, no.~7, pp. 2937--2944, 2018.

\bibitem{paszke2019pytorch}
A.~Paszke, S.~Gross, F.~Massa, A.~Lerer, J.~Bradbury, G.~Chanan, T.~Killeen,
  Z.~Lin, N.~Gimelshein, L.~Antiga \emph{et~al.}, ``Pytorch: An imperative
  style, high-performance deep learning library,'' in \emph{Advances in neural
  information processing systems}, 2019, pp. 8026--8037.

\bibitem{kingma2014adam}
D.~P. Kingma and J.~Ba, ``Adam: A method for stochastic optimization,''
  \emph{arXiv preprint arXiv:1412.6980}, 2014.

\bibitem{zhang2021safe}
Z.~Zhang and J.~F. Fisac, ``Safe occlusion-aware autonomous driving via
  game-theoretic active perception,'' \emph{arXiv preprint arXiv:2105.08169},
  2021.

\bibitem{shu2020autonomous}
K.~Shu, H.~Yu, X.~Chen, L.~Chen, Q.~Wang, L.~Li, and D.~Cao, ``Autonomous
  driving at intersections: a critical-turning-point approach for left turns,''
  in \emph{2020 IEEE 23rd International Conference on Intelligent
  Transportation Systems (ITSC)}.\hskip 1em plus 0.5em minus 0.4em\relax IEEE,
  2020, pp. 1--6.

\end{thebibliography}
}
\end{document}